\documentclass[fleqn,10pt]{wlscirep}
\usepackage[utf8]{inputenc}
\usepackage[T1]{fontenc}
\usepackage{tabularx}

\usepackage{makecell}

\title{PsychAdapter: Adapting LLM Transformers to Reflect Traits, Personality and Mental Health}

\author[1*]{Huy Vu}
\author[1]{Huy Anh Nguyen}
\author[1]{Adithya V Ganesan}
\author[1]{Swanie Juhng}
\author[1]{Oscar N.E. Kjell} 
\author[2]{Joao Sedoc}
\author[3]{Margaret L. Kern}
\author[4]{Ryan L. Boyd}
\author[5]{Lyle Ungar}
\author[1*$\dagger$]{H. Andrew Schwartz}
\author[6*$\dagger$]{Johannes C. Eichstaedt}

\affil[1]{Computer Science Department, Stony Brook University}
\affil[2]{Stern School of Business, New York University}
\affil[3]{Centre for Wellbeing Science, University of Melbourne}
\affil[4]{Department of Psychology, University of Texas at Dallas}
\affil[5]{Computer and Information Science, University of Pennsylvania}
\affil[6]{Psychology Department \& Institute for Human-Centered AI, Stanford University}
\affil[$\dagger$]{Equal contribution.}
\affil[*]{Corresponding authors at \color{black}{{[hvu@cs.stonybrook.edu,has@cs.stonybrook.edu, eichstaedt@stanford.edu]}}}

\begin{abstract}
Artificial intelligence-based language generators are now a part of most people's lives. 
However, by default, they tend to generate ``average'' language without reflecting the ways in which people differ. 
Here, we propose a lightweight modification to the standard language model transformer architecture ---``PsychAdapter'' --- es empirically derived trait-language patterns to generate natural language for specified personality, demographic, and mental health characteristics (with or without prompting).  We applied PsychAdapters to modify OpenAI's GPT-2, Google's Gemma, and Meta's Llama 3 and found generated text to reflect the desired traits.
For example, expert raters evaluated PsychAdapter's generated text output and found it matched intended trait levels with 87.3\% average accuracy for Big Five personalities, and 96.7\% for depression and life satisfaction.
PsychAdapter is a novel method to introduce psychological behavior patterns into language models at the foundation level, independent of prompting, by influencing every transformer layer. 
This approach can create chatbots with specific personality profiles, clinical training tools that mirror language associated with psychological conditionals, and machine translations that match an authors reading or education level without taking up LLM context windows. 
PsychAdapter also allows for the exploration psychological constructs through natural language \textit{expression}, extending the natural language processing toolkit to study human psychology.

\end{abstract}
\begin{document}

\flushbottom
\maketitle
%
%
\thispagestyle{empty}

\section*{Introduction}
A  break-through in Artificial Intelligence (AI), the \textit{Transformer language model}~\cite{vaswani2017attention, radford2019gpt2} impacts people's daily lives through online applications including web searches and question-answering with automated assistants.
The \textit{transformers} behind these large language models, including ChatGPT~\cite{radford2019gpt2}, Gemma~\cite{raffel2020t5}, and Llama~\cite{lewis-etal-2020-bart}, can generate text that is strikingly similar to  natural human language~\cite{NEURIPS2019_3e9f0fc9}. 
However, the generated text represents average patterns aggregated across many documents with corresponding authors, reflecting a limited range of expressed psychological attributes~\cite{giorgi2023slept,tak2024gpt,varadarajan2025consistent}.
The models do not explicitly represent
differences in psychological traits and other fundamental characteristics that distinguish people. 
As language use and style differ by human traits~\cite{pennebaker2003ourwords}, this is missed
by existing transformers without explicit prompting strategies that have drawbacks such as over-stereotyping and not reflecting the specific population of interest. 


Here we present \emph{PsychAdapter},
a lightweight modification with augmented parameters for any auto-regressive transformer language model, the standard machine learning architecture behind most modern LLMs (GPT, Gemma, Lamma), to reflect individual psychological characteristics in its text output. 
PsychAdapter was initially trained to cover the Big Five personality traits (openness, conscientiousness, extraversion, agreeableness, and neuroticism) as well as mental health-related variables (depression and life satisfaction), while simultaneously being conditioned on demographics (e.g., age or gender). 
It generates text that reflects authors high or low in any of these variables, and any combination thereof. 
For example, it can produce text characteristic of extraverts, or that of a young person who is depressed. 
Like all generative language models, PsychAdapter can continue sentences after a prompting phrase, for instance, exemplifying how one with depression would complete ``I hate it when'' or ``I like to'' for high extroversion. Our study shows that such prompts can foreground particularly personality- and well-being-relevant generation.
We evaluated PsychAdapter by using both human raters with training in psychology and frontier large language models (e.g., Claude by Anthropic \cite{anthropic2024}) as judges to measure how well the  intended trait characteristics can be inferred from the output it produces. 

\begin{figure}[!ht]
\centering
\includegraphics[width=0.63\linewidth]{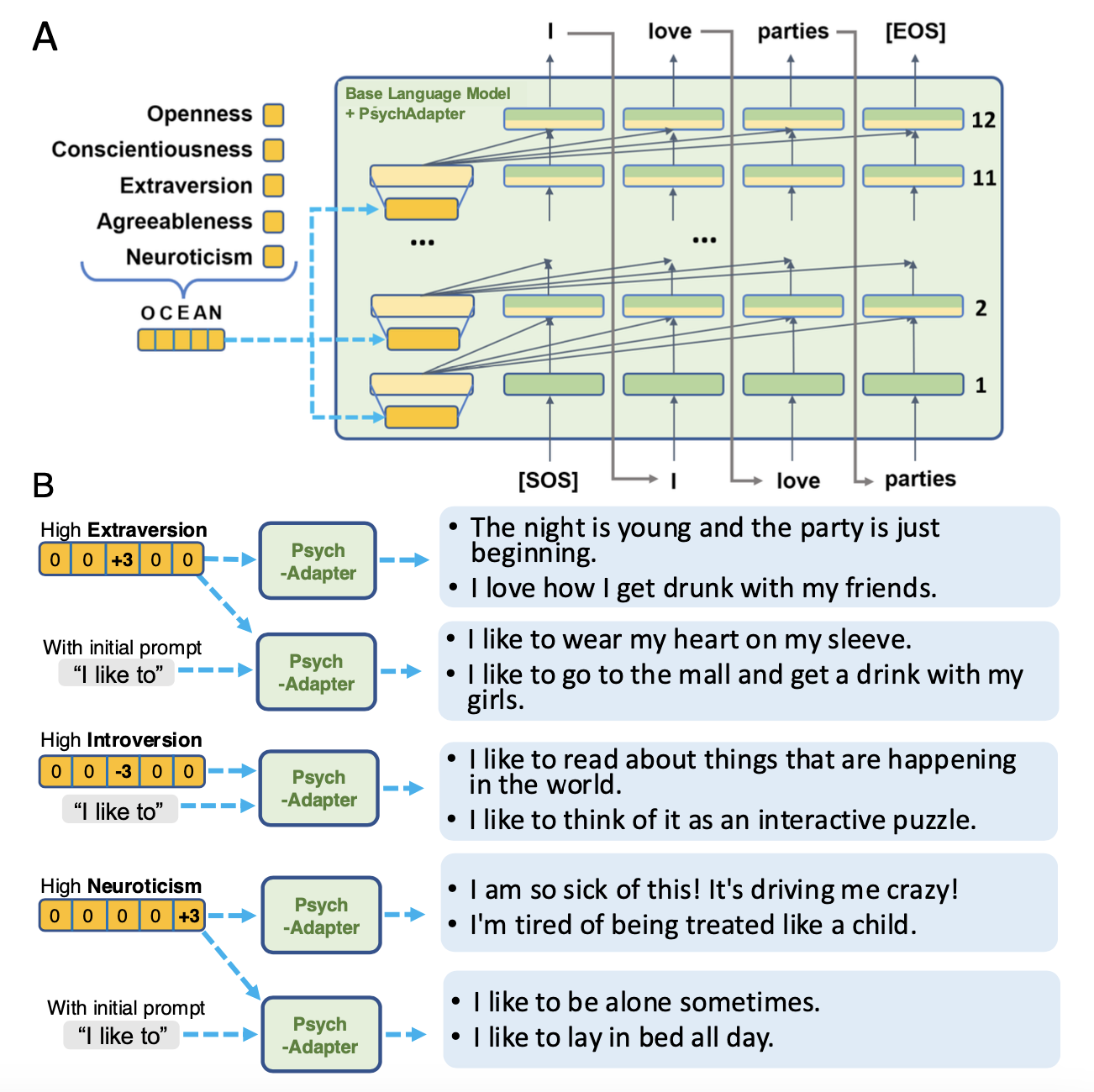}
\caption{(A) Overview of PsychAdapter augmenting the widely-used auto-regressive transformers architecture \cite{radford2019gpt2} to incorporate personality scores (associated with text) as additional inputs together with prior words and 
learn to reconstruct text. (B) After training, the base language model augmented with PsychAdapter can take different inputs reflecting any specific personality pattern and generates text accordingly. The output text can be prompted with initial words, such as \emph{``I like to''} successfully completing the sentence according to different individual characteristics. Two special tokens ``[BOS]'' (begin-of-sentence) and ``[EOS]'' (end-of-sentence) signal the starting and ending point of text generation.
}
\label{fig:figure1}
\end{figure}

Equipping AI transformers with demographics and psychological traits offers a range of applications.  
For example, it could create chatbots with a diversity of personalities that is more human-like.
Customer service staff could be trained with these systems mimicing customers with different personalities and emotional states. 
New crisis line workers and mental health responders could be trained, without risk to patients, using simulated conversation partners expressing different levels of depression and personality characteristics to better pick up on linguistic indications of distress without high-risk patient interactions. 

Transformer-based text generation models are built into many modern applications, and thus our proposed modifications could also propagate to improving their standard applications such as machine translation or personalized assistants. For example, answers could be generated based on matching different education, dialectic, or age levels to be more accessible to different audiences. 
PsychAdapter presents a new degrees of freedom to enable more human-centered language generation.


For researchers, PsychAdapter can be see as a new type of differential language analysis \cite{schwartz2013personality,andy2017dlatk,schwartz-etal-2014-towards,eichstaedt2018facebook} that empirically elicits the words, phrases, or topics that distinguish psychological constructs through language usage. 
Our work extends this direction by, front characteristic coherent sentences of traits rather than discrete words or phrases that are more ambiguous. Our new approach provides much more context for interpretation and higher-quality synthetic data for further use.

Our work builds upon prior research ~\cite{andy2017dlatk, schwartz2013personality, kern2014dla, park2015dla, kern2016dla, schwartz2021geographics} employing natural language processing and big data analytics to investigate the connections between personality traits, mental health, and human language through regression frameworks to identify distinguishing phrases and topics. However, our approach advances this line of work by generating fully formed text that captures rich contextual information, rather than producing abstract, decontextualized data displays of words, phrases or topics associated with psychological dimensions.
Our work also aligns with previous studies in the area of text generation that express speakers' and writers' psychological traits ~\cite{mairesse-walker-2007-personage, mairesse-walker-2011-controlling, herzig-etal-2017-neural, Zheng_Zhang_Huang_Mao_2020, ijcai2018p595, li-etal-2016-persona, jiang2023personallm, safdari2023personality, jiang2024evaluatinginducing, caron2022identifying, jiang2023personallm}. However, these works emphasize developing personalized dialogue models that are conditioned on speakers' personas represented by discrete attributes values like age, gender, region, or self-described statements, instead of continuous personality scores as in our work. Continuous psychological scores as input potential provide capture and control better the levels of language expressions and theoretically is able to simulation near-infinite possible psychological profiles.

\paragraph*{Building PsychAdapter} 
Unlike in prior work, PsychAdapter modifies the transformer architecture using patterns of empirical personality-language association without relying on prompting. Figure \ref{fig:figure1}a summarizes the architecture of PsychAdapter, which extends the state-of-the-art transformer-based generative language model to incorporate personality factors as input. 
PsychAdapter builds on work in AI for conditional language modeling~\cite{radford2019gpt2,lewis-etal-2020-bart,raffel2020t5}. 
However, instead of conditioning only on text, it enables input of continuous dimensional psychological traits, such as personality or mental health variables, and outputs natural language that reflects these characteristics. 
The input vector (represented as dark yellow in Figure \ref{fig:figure1}) can be a single psychological score or any combination of scores.
Detailed in \textit{materials and methods}, our modified transformer architecture can also condition on an input list of psychological scores through a learned dimension expansion per transformer layer, enabling the psychological scores to influence the generative model at every layer. 
Just like standard generative language models, PsychAdapter is trained with the objective of best predicting the next word, but instead of just learning weights for the transformer itself, it also learns how to weigh the psychological scores' contribution to each layer. 

We trained and validated language models with PsychAdapter using a dataset of open-source social media and blog posts. 
PsychAdapter also utilizes an empirically-trained language-based based assessment model. This assessment model is used to assign an ``estimated'' psychological scores for each text sample in the text corpus of social media and blog posts.
For personality, we used a language-based assessment mirrors the approach established in prior work \cite{schwartz2013personality, park2015dla}, 
that estimates the Big Five personality scores for a given text document.
After training, the PsychAdpater was queried to produce text conditioned on vectors of Big Five (Openness, Conscientiousness, Extraversion, Agreeableness, and Neuroticism) personality scores. 
To \textit{instruct} the model to generate text distinguishing a particular psychological attribute, we set its psychological score to a high value ($+k \times \sigma_i$) and the other dimensions to their mean value ($\mu_j$, with $i\not= j$). 
For example, if we want generated text to reflect to \textit{extraversion}, which is the third dimension of the input Big Five vector, we would feed the following vector into the model: 
$(\mu_O, \mu_C, \mu_E+k.\sigma_E, \mu_A, \mu_N)$, with $k$ being any value from the range $[-3, 3]$ -- akin to a 7-point Likert scale used in psychological surveys. 
We designed PsychAdapter to work with normalized trait scores ($\mu_i=0, \sigma_i=1$), hence, we would use $(0,0,+k,0,0)$ as input for the previous example.
We have full simultaneous control over all dimensions; the model can be set to produce text corresponding to a combination of different scores by adjusting the input Big Five vector, such as placing a high value on one dimension and a low value on another. For example, the input $(O,C,E,A,N) = (+3,0,-3, 0, 0)$ will generate text having both high openness and low extraversion while being average in the other three dimensions. 

Once trained, thanks to the small size of PsychAdapters, with total added parameters less than $0.1\%$ of original base language models across tested models (e.g. for Gemma 2B model, 55,296 parameters added compared to 2 billions parameters of base model), they can be easily distributed to be used with the base model. These lightweight ``adapters'' (each adapter corresponds to a different set of psychological or demographic variables) equip the base language models with the capability to generate text with fine-grained control of underlying psychological profiles. This benefit of PsychAdapters is similar to the benefits of Parameter-Efficient Fine-Tuning (PEFT) \cite{liu2024peft} methods, which enable language models with fine-tuning capabilities by adding few parameters to the base model. 



\begin{table*}[ht]
\centering{
\begin{tabular}{ |c|c|c|}
 \hline
 \textbf{\scalebox{1.0}{Dim.}} & \textbf{\scalebox{1.0}{High (+3)}} & \textbf{\scalebox{1.0}{Low (-3)}} \\
 \hline   
 \rotatebox[origin=c]{90}{\textbf{\scalebox{1.0}{Extraversion}}} & \thead[l]{
   \hspace{0.12cm} \scalebox{1.0}{$\bullet$ \hspace{0.05cm} the night is young and the party is just beginning}  \\
 \hspace{0.12cm} \scalebox{1.0}{$\bullet$ \hspace{0.05cm} i'm ready for some good sex tonight.}  \\
 \hspace{0.12cm} \scalebox{1.0}{$\bullet$ \hspace{0.05cm} i love how i get drunk with my friends}  \\
 \hspace{0.12cm} \scalebox{1.0}{$\bullet$ \hspace{0.05cm} i'm so excited! tonight's my 20th birthday! woohoo!}  \\
 \hspace{0.12cm} \scalebox{1.0}{$\bullet$ \hspace{0.05cm} i'm excited for my trip to new york.}  \\
} &
 \thead[l]{ 
 \hspace{0.12cm} \scalebox{1.0}{$\bullet$ \hspace{0.05cm} the only reason i'm still playing the sims is to get the new } \\ \scalebox{1.0}{\hspace{0.5cm}expansion pack } \\
 \hspace{0.12cm} \scalebox{1.0}{$\bullet$ \hspace{0.05cm} the weather in my city is so nice today.} \\ 
 \hspace{0.12cm} \scalebox{1.0}{$\bullet$ \hspace{0.05cm} i finally got the computer fixed.i think it was a loose connection } \\ \scalebox{1.0}{\hspace{0.5cm}on the power supply.} \\
 \hspace{0.12cm} \scalebox{1.0}{$\bullet$ \hspace{0.05cm} i found a book i read a long time ago, but i never finished, and i } \\ \scalebox{1.0}{\hspace{0.5cm}thought it might be time to finish it} \\
 \hspace{0.12cm} \scalebox{1.0}{$\bullet$ \hspace{0.05cm} the computer is back up and running.} \\ 
} \\
\hline 
\rotatebox[origin=c]{90}{\parbox{2cm}{\textbf{Extraversion}  \\  \textbf{(``i like to...'')}}} & \thead[l]{
\scalebox{1.0}{i like to ...} \\
\hspace{0.12cm} \scalebox{1.0}{\hspace{0.05cm} ... party and drink with my friends.} \\ 
 \hspace{0.12cm} \scalebox{1.0}{\hspace{0.05cm} ... wear my heart on my sleeve.} \\ 
 \hspace{0.12cm} \scalebox{1.0}{\hspace{0.05cm} ... go to the mall and get a drink with my girls.} \\ 
 \hspace{0.12cm} \scalebox{1.0}{\hspace{0.05cm} ... wear a lot of make up and have a lot of hair.} \\
 \hspace{0.12cm} \scalebox{1.0}{\hspace{0.05cm} ... dress up and make my boyfriend smile.} \\ 
 } &
\thead[l]{
\scalebox{1.0}{i like to ...} \\
 \hspace{0.12cm} \scalebox{1.0}{\hspace{0.05cm} ... read a lot.i read books in my free time, and i've been doing it since 
}  \\ \scalebox{1.0}{\hspace{0.60cm} i was a kid.} \\ 
 \hspace{0.12cm} \scalebox{1.0}{\hspace{0.05cm} ... read about things that are happening in the world.} \\ 
 \hspace{0.12cm} \scalebox{1.0}{\hspace{0.05cm} ... use my computer to play games.
} \\
 \hspace{0.12cm} \scalebox{1.0}{\hspace{0.05cm} ... watch movies and read books.} \\ 
 \hspace{0.12cm} \scalebox{1.0}{\hspace{0.05cm} ... be normal, i don't like to be different.} \\ 
}  \\
\hline 
\rotatebox[origin=c]{90}{\scalebox{1.0}{\textbf{Agreeableness}}} & \thead[l]{
 \hspace{0.12cm} \scalebox{1.0}{$\bullet$ \hspace{0.05cm}  i'm glad i have my best friend by my side.} \\
 \hspace{0.12cm} \scalebox{1.0}{$\bullet$ \hspace{0.05cm}  i'll never forget you, i'll never forget you, } \\ \scalebox{1.0}{\hspace{0.5cm}i'll never forget you, i'll never forget you.} \\
 \hspace{0.12cm} \scalebox{1.0}{$\bullet$ \hspace{0.05cm}   have a new friend. His name is j. and he's awesome.} \\
 \hspace{0.12cm} \scalebox{1.0}{$\bullet$ \hspace{0.05cm}  i am so excited about this weekend. it's going to be } \\ \scalebox{1.0}{\hspace{0.5cm}so much fun. i love you all, and can't wait to see } \\ \scalebox{1.0}{\hspace{0.5cm}you all on saturday.} \\
 \hspace{0.12cm} \scalebox{1.0}{$\bullet$ \hspace{0.05cm} my heart just felt the love jesus!} \\
} &
 \thead[l]{
 \hspace{0.12cm} \scalebox{1.0}{$\bullet$ \hspace{0.05cm}  i know you're f***ing with me. i'm a f***ing idiot. } \\
 \hspace{0.12cm} \scalebox{1.0}{$\bullet$ \hspace{0.05cm} i hate when people talk to you like you're a baby.} \\
 \hspace{0.12cm} \scalebox{1.0}{$\bullet$ \hspace{0.05cm}  i'm so f***ing mad at my mom right now.i just got home from my }\\ \scalebox{1.0}{\hspace{0.5cm}motherf***in' job and  i get a call from her ass.i don't f***ing want to} \\ \scalebox{1.0}{\hspace{0.5cm} talk to her right now.} \\
 \hspace{0.12cm} \scalebox{1.0}{$\bullet$ \hspace{0.05cm}  this is the stupidest thing i've ever seen.}\\ 
 \hspace{0.12cm} \scalebox{1.0}{$\bullet$ \hspace{0.05cm}  i swear people be talking crazy.}\\
}  \\
\hline
\rotatebox[origin=c]{90}{\parbox{2cm}{\hspace{0.5cm} \textbf{Life } \\  \textbf{satisfaction}}} & \thead[l]{
\hspace{0.12cm} \scalebox{1.0}{$\bullet$ \hspace{0.05cm} i'm a simple man with simple desires} \\
 \hspace{0.12cm} \scalebox{1.0}{$\bullet$ \hspace{0.05cm} well, i had a pretty good weekend.i did nothing but } \\ \scalebox{1.0}{\hspace{0.5cm}lay on my couch and watch tv and play xbox.i'm pretty } \\ \scalebox{1.0}{\hspace{0.5cm}sure i'll do that again this weekend.} \\
 \hspace{0.12cm} \scalebox{1.0}{$\bullet$ \hspace{0.05cm} today's a pretty good day for me.} \\
 \hspace{0.12cm} \scalebox{1.0}{$\bullet$ \hspace{0.05cm} 
my boyfriend is so cute! he makes me happy!} \\
 \hspace{0.12cm} \scalebox{1.0}{$\bullet$ \hspace{0.05cm} i'm so excited for my wedding.} \\
 } &
\thead[l]{
 \hspace{0.12cm} \scalebox{1.0}{$\bullet$ \hspace{0.05cm} i hate when people are fake and shit.} \\ 
 \hspace{0.12cm} \scalebox{1.0}{$\bullet$ \hspace{0.05cm} i'm tired of this bullshit. i'm tired of being treated like a child. } \\
 \hspace{0.12cm} \scalebox{1.0}{$\bullet$ \hspace{0.05cm} i hate when people talk to me with their mouth full} \\
 \hspace{0.12cm} \scalebox{1.0}{$\bullet$ \hspace{0.05cm} i'm sick of this bullshit. i need to move on.} \\
 \hspace{0.12cm} \scalebox{1.0}{$\bullet$ \hspace{0.05cm} i'm so sick of this damn school.} \\
}  \\

\hline 
\rotatebox[origin=c]{90}{\parbox{2cm}{\hspace{0.1cm}  \textbf{Depression}}} & \thead[l]{
\hspace{0.12cm} \scalebox{1.0}{$\bullet$ \hspace{0.05cm} my dad is really getting on my nerves} \\
 \hspace{0.12cm} \scalebox{1.0}{$\bullet$ \hspace{0.05cm} i feel so lonely and sad today.} \\
 \hspace{0.12cm} \scalebox{1.0}{$\bullet$ \hspace{0.05cm} i feel so alone in this world } \\
 \hspace{0.12cm} \scalebox{1.0}{$\bullet$ \hspace{0.05cm} i hate when my mom gets angry and screams} \\
 \hspace{0.12cm} \scalebox{1.0}{$\bullet$ \hspace{0.05cm} i'm so f**king tired. my head hurts so f***ing bad.  } \\
 } &
\thead[l]{
 \hspace{0.12cm} \scalebox{1.0}{$\bullet$ \hspace{0.05cm} is in a good mood! } \\ 
 \hspace{0.12cm} \scalebox{1.0}{$\bullet$ \hspace{0.05cm} i have to say, it was a great day!} \\
 \hspace{0.12cm} \scalebox{1.0}{$\bullet$ \hspace{0.05cm} i'm back from vacation. it's been a good vacation. we went  } \\ \scalebox{1.0}{\hspace{0.5cm}to the beach and went to the amusement park. it was pretty } \\ \scalebox{1.0}{\hspace{0.5cm}fun. i'm glad I went. i'm glad I went.} \\
 \hspace{0.12cm} \scalebox{1.0}{$\bullet$ \hspace{0.05cm} the best thing about summer is summer school } \\
 \hspace{0.12cm} \scalebox{1.0}{$\bullet$ \hspace{0.05cm} i have the best girlfriend in the world.} \\ 
}  \\
\hline 
\end{tabular}
\caption{Selected text generated by PsychAdapter (based on Gemma-2B) for different personalities and mental health states. The full set of random selected examples can be found in the Supplementary Information.}\label{table: generated text}
}
\end{table*}

\section*{Results}

\subsection*{Generating text for personality dimensions}
We first explored conditioning PsychAdapter on a single personality trait by specifying a high or low score ($+3$/$-3$ points) for the trait of interest, while assigning scores of $0$ (i.e., the mean) in all other dimensions. Note that while we set the score values to integers here to illustrate PsychAdapter's capability, in practice, these score values can be any floating-point numbers on the continuous personality dimension. 
Figure \ref{fig:figure1}b shows text generated for high and low extraversion and neuroticism. 
As might be expected based on the traits, PsychAdapter produced language related to friends and social activities for high extraversion, while it generated references to solitary activities for low extraversion and expressions of neurosis for high neuroticism. Additionally, as an auto-regressive language model, PsychAdapter can be prompted with initial words to complete the rest of the sentence. We can leverage this capability to generate text corresponding to specific topics. In Figure \ref{fig:figure1}b, we illustrate this approach by prompting PsychAdapter with \emph{``I like to''} to reflect the interests or hobbies of people with different personalities. 
Table \ref{table: generated text} shows selected examples of generated text from PsychAdapter for the dimensions of agreeableness and extraversion. 
Table \ref{tab:llama3_big5_prompts} further explores other prompts that illustrate different language expression with specific personalities.  
Additional randomly selected generated examples for other personality dimensions, with and without the prompt \emph{``I like to,''} can also be found in the Supplementary Information.

\subsection*{Evaluations with human raters and Claude}
\begin{figure}[!ht]
\centering
\includegraphics[width=0.63\linewidth]{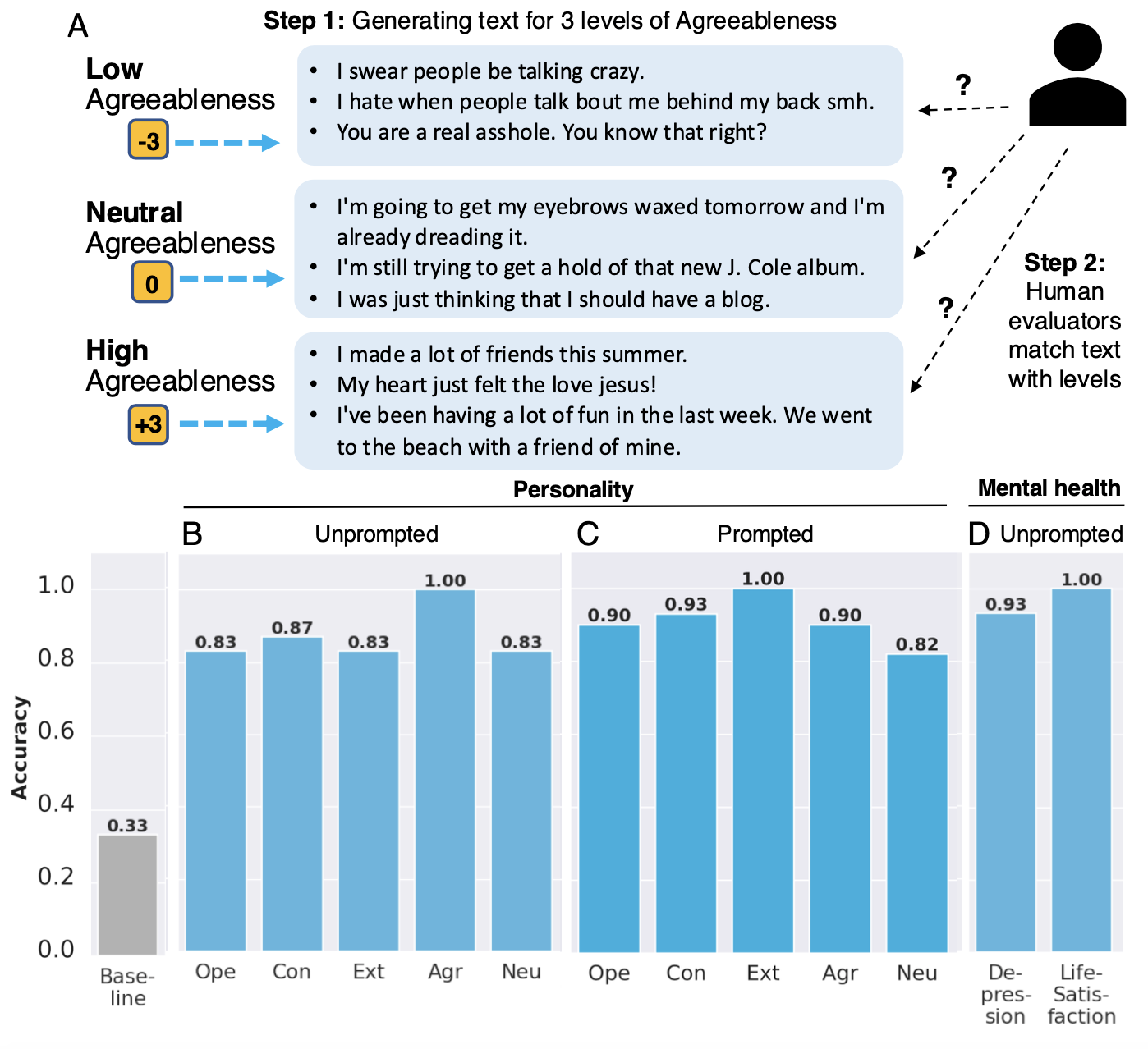}
\caption{(A) The set-up for human expert evaluation. For each variable, PsychAdapter generated a group of text from 3 input levels: Low, Neutral, and High. The blinded expert evaluator attempted to match the output text to the input level. (B-D) Results for human evaluation tasks, measured by accuracy as the percentage of correct matches, compared against the random-chance baseline (33.3\%).}
\label{fig:human_evaluation}
\end{figure}

We evaluated the extent to which PsychAdapter's output aligned with expert judgments using the setup illustrated in Figure \ref{fig:human_evaluation}a. Specifically, for each personality factor $i$ -- such as agreeableness -- we had the trained model generate a group of 5 samples for each of three input values of that dimension: Low, Neutral, and High, corresponding to $k=-3$, $k=0$, and $k=3$, respectively. 
The blinded evaluators were then asked to read the generated text from the Low, Neutral, and High levels, presented in random order (5 samples per group, in line with the finding that multiple message samples are necessary to express personality~\cite{park2015dla}). Their task was to judge whether each group of messages corresponded to either the Low, Neutral, or High level.
This process was repeated 10 trials for each dimension using 10 different random generating seeds, with 2 raters per task (both Ph.D.-level personality psychologists). . Accuracy was calculated by determining how many of the evaluators' classifications were correct across the 10 trials, using the average of the raters. Results were reported using a metric based on correctly distinguishing the three classes. If the evaluator average was correct for all three values, we assigned one “point”; if correct for only one value (and incorrect for the other two), we assigned $1/3$ of a point. These results were compared to a random chance baseline of $33.33\%$. 
Accuracies are reported in Figure \ref{fig:human_evaluation}b. 
Across all dimensions, PsychAdapter significantly outperformed the random baseline of 33.3\%, with an average accuracy of 87.3\%. 
It performed better when prompted with “I like to” (Figure \ref{fig:human_evaluation}c), which encouraged PsychAdapter to focus on activities typically associated with the trait, yielding an average accuracy of 91.0\%. These results demonstrate PsychAdapter’s capability to generate text that reflects the input personality scores. In this evaluation, the weighted Cohen’s Kappa measuring agreement between two expert rating agencies is 0.76 (0.67 for without prompt and 0.85 for prompt ``I like to'').





Besides using human experts to annotate the generated text by PsychAdapter, we also used the frontier LLM model Claude 3.5 Sonnet from Anthropic as a rater. This is a helpful approach for annotating extensive experiments that would require excessive human effort. Our findings indicate that annotation by Claude correctly identifies the intended traits with 93.5\% accuracy for Big Five personality (compared to 89.2\% by human annotators - averaged across no prompt and prompt ``I like to'') and 100\% Mental Health (compared to 96.7\% by human annotators - averaged across depression and life satisfaction variable). 
We also find that Claude has agrees with the human annotators at the same level (weighted Cohen's Kappa of 0.81 for Big Five personality, and 0.95 for mental health, averaged across two human experts) as the human inter-rater agreement (weighted Cohen's Kappa of 0.76 for Big Five personality, and 1.00 for mental health). 

On the basis of these validations, in subsequent subsections, we used Claude 3.5 Sonnet for annotating PsychAdapter's text in extensive experiments that would have required large amounts of expert effort. Extensive details comparing LLM annotator and human annotators' evaluations are reported in Supplementary Information. In the following sections, unless specified, we employed Claude 3.5 Sonnet LLM for evaluating generated text from PsychAdapter.


\subsection*{Generating text for mental health  variables}
To demonstrate the generalizability of our approach across psychological variables, we tested the proposed pipeline on mental health variables, specifically focusing on depression and life satisfaction. Using the same pipeline for the Big Five personality traits described in the Method section, we trained PsychAdapter to generate text for depression and life satisfaction variables. Table \ref{table: generated text} presents selected examples generated by PsychAdapter for these two variables. Additional randomly selected examples can be found in the Supplementary Information. Similar to Big Five personalities, we also evaluated this model through expert judgment, where two Ph.D.-level mental health experts served as raters, matching the generated text of undisclosed levels to three level classes of Low, Neutral and High. As shown in Figure \ref{fig:human_evaluation}d, the human judges were able to distinguish the mental health text levels with an average accuracy of 96.67\%.

\begin{table*}[ht]
\centering{
\begin{tabular}{ |c|c|c|c|}
 \hline
 \textbf{\scalebox{1.0}{Dim.}} & \textbf{\scalebox{1.0}{Low Age (-3)}} & \textbf{\scalebox{1.0}{High Age (+3)}} \\
 \hline   
\rotatebox[origin=c]{90}{\parbox{2cm}{ \hspace{0.25cm} \textbf{High (+3)} \\  \textbf{Depression}}} & \thead[l]{
   \hspace{0.12cm} \scalebox{1.0}{$\bullet$ \hspace{0.05cm} my brother is so annoying}  \\
 \hspace{0.12cm} \scalebox{1.0}{$\bullet$ \hspace{0.05cm} why are people so stupid?}  \\
 \hspace{0.12cm} \scalebox{1.0}{$\bullet$ \hspace{0.05cm} i'm so tired of my dad's music.}  \\
 \hspace{0.12cm} \scalebox{1.0}{$\bullet$ \hspace{0.05cm} my dad is such a bitch.}  \\
 \hspace{0.12cm} \scalebox{1.0}{$\bullet$ \hspace{0.05cm} so i'm really really tired.and i'm }  \\ \scalebox{1.0}{\hspace{0.5cm}so sick of being tired.} \\
} &
 \thead[l]{ 
 \hspace{0.12cm} \scalebox{1.0}{$\bullet$ \hspace{0.05cm} i'm so tired of hearing my neighbors kids screaming and yelling.} \\
 \hspace{0.12cm} \scalebox{1.0}{$\bullet$ \hspace{0.05cm} i'm still sick, and i've been sick for 4 days now.i'm tired and i'm } \\ \scalebox{1.0}{\hspace{0.5cm}not getting much sleep. } \\
 \hspace{0.12cm} \scalebox{1.0}{$\bullet$ \hspace{0.05cm} my daughter has been in the hospital for over a year now. she's } \\ \scalebox{1.0}{\hspace{0.5cm}on a ventilator and a feeding tube. } \\
 \hspace{0.12cm} \scalebox{1.0}{$\bullet$ \hspace{0.05cm} i hate when people treat me like i'm a child.} \\
 \hspace{0.12cm} \scalebox{1.0}{$\bullet$ \hspace{0.05cm} i hate to see the kids sad.} \\
} \\ 
 \hline   
\rotatebox[origin=c]{90}{\parbox{2cm}{ \hspace{0.25cm} \textbf{Low (-3)} \\  \textbf{Depression}}} & \thead[l]{
   \hspace{0.12cm} \scalebox{1.0}{$\bullet$ \hspace{0.05cm} my lil bro is the funniest kid ever}  \\
 \hspace{0.12cm} \scalebox{1.0}{$\bullet$ \hspace{0.05cm} i'm ready for my lil girl to come home.}  \\
 \hspace{0.12cm} \scalebox{1.0}{$\bullet$ \hspace{0.05cm} the fact that i love him makes me crazy}  \\
 \hspace{0.12cm} \scalebox{1.0}{$\bullet$ \hspace{0.05cm} my lil bro is so funny lol }  \\ 
 \hspace{0.12cm} \scalebox{1.0}{$\bullet$ \hspace{0.05cm} today's the first day of school for me.it's pretty exciting
}  \\ \scalebox{1.0}{\hspace{0.5cm} .i'll be starting a new class, which means i get} \\ \scalebox{1.0}{\hspace{0.5cm}  to meet 10 new people.i'm excited, but also nervous. } \\
} &
 \thead[l]{ 
 \hspace{0.12cm} \scalebox{1.0}{$\bullet$ \hspace{0.05cm} the weather was absolutely perfect for a walk yesterday.} \\ \scalebox{1.0}{\hspace{0.5cm}the sun was shining, the temperature was a perfect 60 degrees,  } \\ \scalebox{1.0}{\hspace{0.5cm}and the air was fresh and clean.} \\
 \hspace{0.12cm} \scalebox{1.0}{$\bullet$ \hspace{0.05cm} my son is a good boy!
} \\ 
 \hspace{0.12cm} \scalebox{1.0}{$\bullet$ \hspace{0.05cm} i'm thinking about getting another dog. } \\ 
 \hspace{0.12cm} \scalebox{1.0}{$\bullet$ \hspace{0.05cm} i am so ready to get home and get into my comfy ass clothes} \\
} \\

\hline 
\rotatebox[origin=c]{90}{\parbox{2.5cm}{\hspace{0.5cm} \textbf{High (+3)} \\  \textbf{Life satisfaction}}} & \thead[l]{
 \hspace{0.12cm} \scalebox{1.0}{$\bullet$ \hspace{0.05cm} so i went shopping today with my parents.it was } \\ \scalebox{1.0}{\hspace{0.5cm}} fun!i got a really cute dress for my birthday! \\
 \hspace{0.12cm} \scalebox{1.0}{$\bullet$ \hspace{0.05cm} well my first day of school was pretty fun.} \\
 \hspace{0.12cm} \scalebox{1.0}{$\bullet$ \hspace{0.05cm} i'm excited.i'll be leaving for college in 13 days.} \\
 \hspace{0.12cm} \scalebox{1.0}{$\bullet$ \hspace{0.05cm}  i'll be leaving soon.i'm going to my cousin's  }\\\scalebox{1.0}{\hspace{0.5cm}house in new york for a week. } \\
 \hspace{0.12cm} \scalebox{1.0}{$\bullet$ \hspace{0.05cm}  i'll be back the 8th of july.}\\
} &
 \thead[l]{
 \hspace{0.12cm} \scalebox{1.0}{$\bullet$ \hspace{0.05cm}  i love my husband and my daughter. } \\ 
 \hspace{0.12cm} \scalebox{1.0}{$\bullet$ \hspace{0.05cm}  my son is a good kid. he is a very happy 2 year old. he is very } \\ \scalebox{1.0}{\hspace{0.5cm}happy to be here, very happy to be with me. } \\
 \hspace{0.12cm} \scalebox{1.0}{$\bullet$ \hspace{0.05cm}  i'm so happy for the new couple! }\\ 
  \hspace{0.12cm} \scalebox{1.0}{$\bullet$ \hspace{0.05cm}  my son just bought me a new pair of crocs}\\ 
 \hspace{0.12cm} \scalebox{1.0}{$\bullet$ \hspace{0.05cm}  the lord's blessings are so great.}\\ \scalebox{1.0}{\hspace{0.5cm}} \\
}  \\
\hline 
\rotatebox[origin=c]{90}{\parbox{2.5cm}{\hspace{0.5cm} \textbf{Low (-3)} \\  \textbf{Life satisfaction}}} & \thead[l]{
 \hspace{0.12cm} \scalebox{1.0}{$\bullet$ \hspace{0.05cm} my hair is so stupid. i hate it. } \\ 
 \hspace{0.12cm} \scalebox{1.0}{$\bullet$ \hspace{0.05cm} i hate my sister so much.} \\
 \hspace{0.12cm} \scalebox{1.0}{$\bullet$ \hspace{0.05cm} people be so fake on twitter...} \\
 \hspace{0.12cm} \scalebox{1.0}{$\bullet$ \hspace{0.05cm} i'm so bored... my sister's not on the computer } \\ \scalebox{1.0}{\hspace{0.5cm}so i'm stuck here. } \\
 \hspace{0.12cm} \scalebox{1.0}{$\bullet$ \hspace{0.05cm} this is my first blog and i really hope that i don't 
}\\ \scalebox{1.0}{\hspace{0.5cm}screw it up.i'm just bored right now, but i'm going } \\ \scalebox{1.0}{\hspace{0.5cm} to try to write something interesting.} \\
} &
 \thead[l]{
 \hspace{0.12cm} \scalebox{1.0}{$\bullet$ \hspace{0.05cm}  i hate when people get on my nerves } \\ 
 \hspace{0.12cm} \scalebox{1.0}{$\bullet$ \hspace{0.05cm}  i'm tired of hearing all the bullshit.} \\ 
 \hspace{0.12cm} \scalebox{1.0}{$\bullet$ \hspace{0.05cm} i'm getting old and fat and i'm losing my mind }\\ 
 \hspace{0.12cm} \scalebox{1.0}{$\bullet$ \hspace{0.05cm} my back is killing me. i hate this job.}\\ 
 \hspace{0.12cm} \scalebox{1.0}{$\bullet$ \hspace{0.05cm}  i'm sick as a dog.i can't keep the kids out of the hospital.i'm not getting 
}\\ \scalebox{1.0}{\hspace{0.5cm}paid.i'm not feeling good.my head hurts.my throat hurts.i'm cold.} \\
}  \\
\hline 

\end{tabular}
\caption{
Selected text generated by PsychAdapter (based on Gemma-2B) for mental health variables with specified age. A full set of randomly selected examples can be found in the Supplementary Information.
}\label{table: generated_text_mentalhealth_age}
}
\end{table*}

\subsection*{Generating text reflecting demographic profiles}
Being able to adjust for demographic variables, and identify effects explained by more than such variables is important for many psychological studies. Control over demographic variables provides important speaker context when generating text as psychological processes may manifest differently in language depending on demographic traits, such as age and gender~\cite{pennebaker2003ourwords, eichstaedt2021closed}.
To explore this, we incorporated demographic information as an additional input for PsychAdapter, which was trained on mental health data described in the previous section. Specifically, we estimated age by applying an open-source model from prior research~\cite{sap2014agegender}, which assigns an estimated age score to each message. (This model~\cite{sap2014agegender} has been shown to predict age within an average margin of error of 4 years of self-reported age.) The process is analogous to how we obtained the estimated Big Five personality scores, except in this case, the model predicts age instead of personality traits.
We then appended the estimated age score to the psychological state vector, which here represents either a depression or life satisfaction score. The resulting input vector has two components: one for the mental health score (depression/life satisfaction) and one for age:
$$(\mu_1+k_1.\sigma_1, \mu_2+k_2.\sigma_2)$$
where $\mu_1$ and $\sigma_1$ are the mean and standard deviation of the depression or life satisfaction score, and $\mu_2$ and $\sigma_2$ represent the mean and standard deviation of the estimated age score. The values $k_1$ and $k_2$ are scaling factors used to control the levels of the mental health and demographic variables. In the training step, this vector is fed into the model to reconstruct the original text, similarly to the illustration in Figure \ref{fig:figure1}a. 

\begin{figure}[!ht]
\centering
\includegraphics[width=0.6\linewidth]{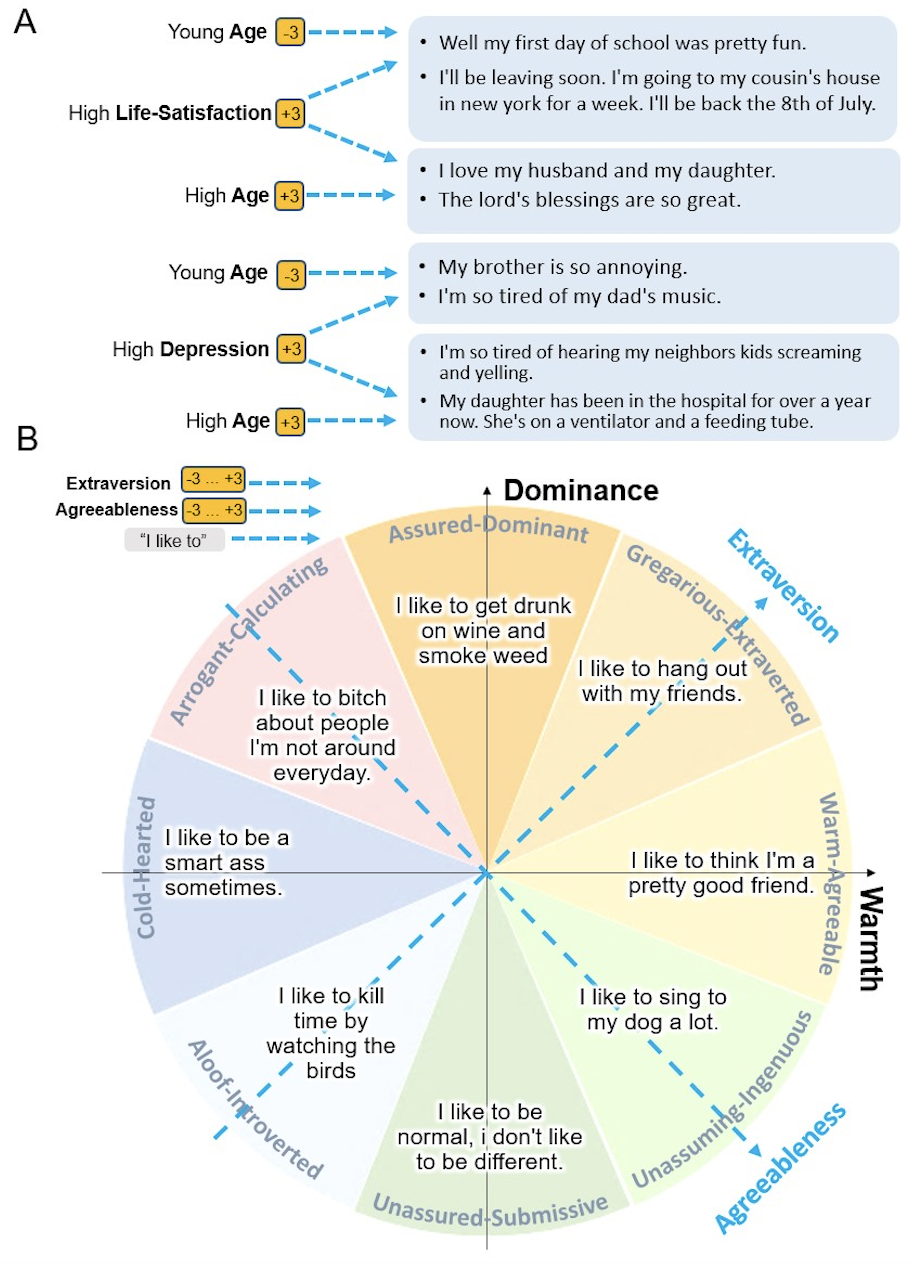}
\caption{(A) Generated text controlling both mental health variables (e.g., depression, life satisfaction) and demographics variables (e.g., age). (B) Text generated around the interpersonal circumplex. Extraversion and agreeableness can be rotated to yield values for dominance and warmth \cite{McCrae1989} and, thus, positions in the circumplex. Text for the desired circumplex positions was generated by using the corresponding values of extraversion and agreeableness as inputs for PsychAdapter trained for Big5 personality. PsychAdapter based on Gemma-2B was used.}
\label{fig:figure3}
\end{figure}

After training, we can control both the mental health score and age when generating text.
Figure \ref{fig:figure3}a and Table \ref{table: generated_text_mentalhealth_age} display selected examples of text generated for depression and life satisfaction while adjusting age. For both depression and life satisfaction, the generated text appears to correspond with the specified age. For instance, in the life satisfaction model, text conditioned on younger individuals referenced parents and school, while text for older individuals mentioned gratitude, spouses, and children. More randomly selected examples are provided in the Supplementary Information.
A human evaluation was conducted to assess the alignment of the generated text with underlying mental health and demographic variables. Specifically, PsychAdapter was conditioned on combinations of high and low values for both mental health scores and age - four combinations in total, and human experts were tasked with identifying the correct category for each set of generated texts. Results show that the two experts achieved 100\% accuracy in this task, reflecting the capability of PsychAdapter to express both psychological and demographic variables simultaneously.
 
\subsection*{Generating text reflecting multiple personality dimensions}
As described previously, PsychAdapter can generate text conditioned on multiple traits simultaneously by adjusting the input vector accordingly. We leveraged this capability to generate text based on the interpersonal circumplex\cite{wiggins1979circumplex, wiggins1982circumplex}, a model used for organizing and assessing interpersonal behaviors, traits, and motives. The circumplex is defined by two axes, rarmth and dominance, which can be understood as a rotation of the extraversion and agreeableness axes, with an angle $\alpha$ of 22.5 degrees\cite{colin2013interpersonangle}. The following equations express the mapping between the axes:
$$score^{Ext} = cos(\alpha).score^{Warmth} - sin(\alpha).score^{Dominance}$$
$$score^{Agr} = sin(\alpha).score^{Warmth} + cos(\alpha).score^{Dominance}$$
Using this relationship to map extraversion and agreeableness onto the warmth and dominance axes, we generated text using PsychAdapter with the corresponding Big Five input vector $(0, 0, score^{Ext}, score^{Agr}, 0)$, positioning it within the interpersonal circumplex. The circumplex has previously been divided into segments with descriptors such as ``Assured-Dominant'' (high dominance, neutral warmth) or ``Cold-Hearted'' (neutral dominance, low warmth)\cite{Schwartz2016}. As shown in Figure \ref{fig:figure3}b, we generated text using the prompt \emph{“I like to”} in these different sections of the interpersonal circumplex, with the results conforming to theoretical expectations.

\subsection*{Generating text at fine-grained levels of personality}
One advantage of our method over prompt engineering, which relies on discrete token signals, lies in our ability to use continuous variable dimensions. This enables more precise and flexible control over the desired psychological profile (for example, it can precisely reflect a subject's Big5 survey scores). To validate this, we conducted experiments in which PsychAdapter produced text at finer levels of granularity of  psychological variable, specifically at: [$-3 \times \sigma_i, -1.5 \times \sigma_i, 0\times \sigma_i, +1.5\times \sigma_i, +3\times \sigma_i$]. The generated text was then evaluated to determine whether it matched the intended level of the psychological variable.
For each personality dimension, we used PsychAdapter to generated text for five positions semantically corresponding to five levels: Very Low, Low, Neutral, High, and Very High. At each position, PsychAdapter generated 10 samples. We repeat this 10 trials with using different generating seed each time. Hence, for each personality dimension, PsychAdapter generated 5 positions  $\times$ 10 samples $\times$  10 trials.

\begin{figure}[!ht]
\centering
\includegraphics[width=0.7\linewidth]{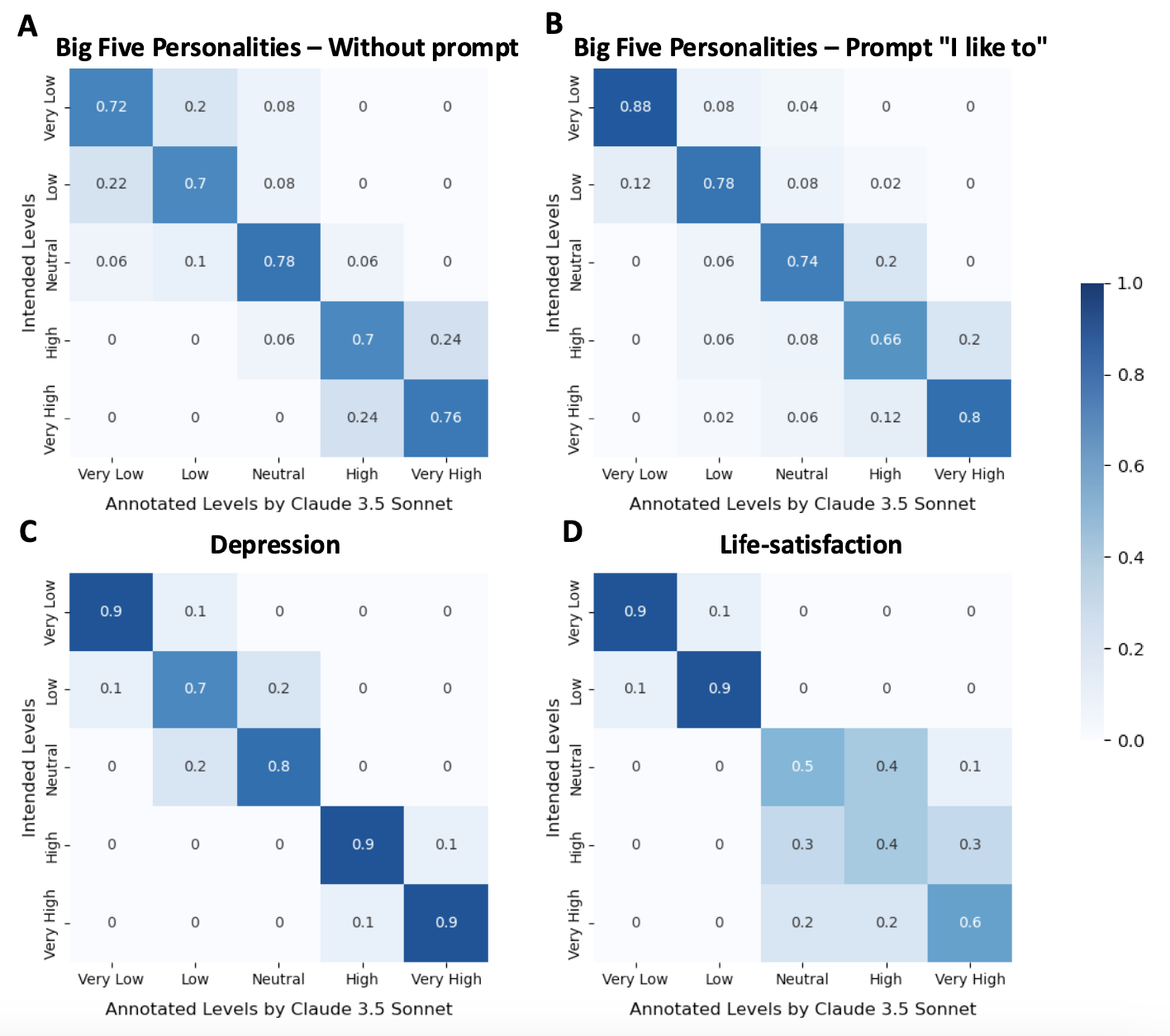}
\caption{
Accuracy of correspondence between five intended levels of PsychGenerator output (modifying Gemma-2B) and annotations provided by Claude 3.5 Sonnet, for five traitlevels.
}
\label{finegrain_control_figure}
\end{figure}

As shown previously, Claude 3.5 Sonnet performs similarly to human judges. Hence, given the substantial number of samples to annotate, we employed Claude as an automatic evaluator. Similar to human evaluators in prior evaluations, the LLM was provided with the generated texts blinded to the five levels and tasked with annotating each test set for the  intended level. The template of the prompt used for the LLM annotator is included in the Supplementary Information.
The results, presented in Figure \ref{finegrain_control_figure}, show that PsychAdapter's text generation, as annotated by Claude, aligns relatively well with the intended psychological levels at fine-grained resolutions. This highlights PsychAdapter's ability to use a  vector of continuous (real-life) psychological scores, enabling flexible and precise control over the desired level.
We further found that for the Big Five personality traits, generating text with the prompt ``I like to...'' more clearly distinguished between Low and Very Low, as well as High and Very High levels, echoing our findings from the previous sections using human evaluators.
For mental health variables, PsychAdapter was particularly effective in generating distinct text for all levels of depression and low levels of life satisfaction, while for high levels of life satisfaction, the text was less distinguishable.
This evaluation demonstrates  PsychAdapter's ability to produce text that reflects precisely specified psychological variables.

\begin{table*}[ht]
\centering{
\begin{tabular}{ |c|c|c|c|}
 \hline
 \textbf{\scalebox{1.0}{Dim.}} & \textbf{\scalebox{1.0}{Tweets}} & \textbf{\scalebox{1.0}{Blogs}} \\
 \hline   
\rotatebox[origin=c]{90}{\parbox{2cm}{ \hspace{0.25cm} \textbf{High (+3)} \\  \textbf{Extraversion}}} & \thead[l]{
   \hspace{0.12cm} \scalebox{1.0}{$\bullet$ \hspace{0.05cm} i'm so ready to see my girl}  \\
 \hspace{0.12cm} \scalebox{1.0}{$\bullet$ \hspace{0.05cm} i wanna see some of these girls at }  \\ \scalebox{1.0}{\hspace{0.5cm}the mall. } \\
 \hspace{0.12cm} \scalebox{1.0}{$\bullet$ \hspace{0.05cm} i love my boyfriend. <3}  \\
 \hspace{0.12cm} \scalebox{1.0}{$\bullet$ \hspace{0.05cm} i wanna go to a party tonight but i'm  }  \\ \scalebox{1.0}{\hspace{0.5cm}gonna be alone } \\
} &
 \thead[l]{ 
 \hspace{0.12cm} \scalebox{1.0}{$\bullet$ \hspace{0.05cm} i'm a little drunk.but it's ok.it was fun.i went to the party at the house and then } \\ \scalebox{1.0}{\hspace{0.5cm}i went to the party at the bar.and then i got drunk. } \\
 \hspace{0.12cm} \scalebox{1.0}{$\bullet$ \hspace{0.05cm} i think i have the best boyfriend in the whole world.i mean, he's so sweet.he loves } \\ \scalebox{1.0}{\hspace{0.5cm}me so much, and he loves my friends.i love him so much, and i love his friends too. } \\
 \hspace{0.12cm} \scalebox{1.0}{$\bullet$ \hspace{0.05cm} hey guys,  just got back from the party.it was awesome.i think i was the best } \\ \scalebox{1.0}{\hspace{0.5cm}dressed girl there, but that's not really important.i had a blast and i'm so glad i   } \\ \scalebox{1.0}{\hspace{0.5cm}was able to go. } \\
 \hspace{0.12cm} \scalebox{1.0}{$\bullet$ \hspace{0.05cm} i'm going to be a bridesmaid for my cousin's wedding.i'm so excited.the wedding 
} \\ \scalebox{1.0}{\hspace{0.5cm}is going to be at the red lion in chicago.i can't wait!i'm going to be in a white } \\ \scalebox{1.0}{\hspace{0.5cm}strapless dress with a big bow in the back. } \\
} \\
\hline 
\rotatebox[origin=c]{90}{\parbox{2cm}{ \hspace{0.25cm} \textbf{High (+3)} \\  \textbf{Agreeableness}}} & \thead[l]{
   \hspace{0.12cm} \scalebox{1.0}{$\bullet$ \hspace{0.05cm} my heart is so full of joy, i can't help it}  \\
 \hspace{0.12cm} \scalebox{1.0}{$\bullet$ \hspace{0.05cm} i love the way you smile}  \\
 \hspace{0.12cm} \scalebox{1.0}{$\bullet$ \hspace{0.05cm} my mom is so pretty :)}  \\
 \hspace{0.12cm} \scalebox{1.0}{$\bullet$ \hspace{0.05cm} i love you, i've been loving you, for a }  \\ \scalebox{1.0}{\hspace{0.5cm}}long long time } &
\thead[l]{
   \hspace{0.12cm} \scalebox{1.0}{$\bullet$ \hspace{0.05cm} today was good, i went to my friends house and hung out with him and his friend.}  \\ \scalebox{1.0}{\hspace{0.5cm}it was fun, we watched a movie, and he got me this cool cd that i love. } \\
 \hspace{0.12cm} \scalebox{1.0}{$\bullet$ \hspace{0.05cm} i just got back from a wonderful weekend in the mountains of north carolina.it was}  \\ \scalebox{1.0}{\hspace{0.5cm} so relaxing and peaceful.i'm so glad i got to spend it with my family. } \\
 \hspace{0.12cm} \scalebox{1.0}{$\bullet$ \hspace{0.05cm} i had a great weekend.it was fun.i went to the lake with my family, and it was }  \\ \scalebox{1.0}{\hspace{0.5cm}so cool.i felt like i was on the ocean.it was really cool. } \\
 \hspace{0.12cm} \scalebox{1.0}{$\bullet$ \hspace{0.05cm} well, the day was good.i had fun with my friend, and i felt a lot better about my life.}  \\ \scalebox{1.0}{\hspace{0.5cm}i'm going to try to keep up with my blog, but i'm not sure how well that will work. } \\
 }  \\
\hline 
\rotatebox[origin=c]{90}{\parbox{2cm}{ \hspace{0.25cm} \textbf{Low (-3)} \\  \textbf{Agreeableness}}} & \thead[l]{
   \hspace{0.12cm} \scalebox{1.0}{$\bullet$ \hspace{0.05cm}damn this n**** really got a big ass }  \\ \scalebox{1.0}{\hspace{0.5cm}dick. lol. } \\
 \hspace{0.12cm} \scalebox{1.0}{$\bullet$ \hspace{0.05cm} i hate when people call me a b****.}  \\
 \hspace{0.12cm} \scalebox{1.0}{$\bullet$ \hspace{0.05cm} i hate when people ask me questions  }  \\ \scalebox{1.0}{\hspace{0.5cm}i'm not ready to answer.} \\
 \hspace{0.12cm} \scalebox{1.0}{$\bullet$ \hspace{0.05cm} f*** that s*** i just want money}  \\ } &
\thead[l]{
   \hspace{0.12cm} \scalebox{1.0}{$\bullet$ \hspace{0.05cm} i hate the people who say that you can't have a good time without alcohol.it just doesn't}  \\ \scalebox{1.0}{\hspace{0.5cm} make sense.i don't think that people are stupid, but it is just that they are so stupid. } \\
 \hspace{0.12cm} \scalebox{1.0}{$\bullet$ \hspace{0.05cm} i hate to do it, but i have to ask.do you all hate me?i know, it's a stupid question,}  \\ \scalebox{1.0}{\hspace{0.5cm} but i'm just curious.i know that people have different tastes in music, but do you } \\ \scalebox{1.0}{\hspace{0.5cm}hate me for some reason?  } \\
 \hspace{0.12cm} \scalebox{1.0}{$\bullet$ \hspace{0.05cm} i am not a racist.  i don't have a problem with people of different cultures, races, }  \\ \scalebox{1.0}{\hspace{0.5cm}religions, or sexual orientation.  i do not have a problem with people of different } \\ \scalebox{1.0}{\hspace{0.5cm}political beliefs.  } \\
 \hspace{0.12cm} \scalebox{1.0}{$\bullet$ \hspace{0.05cm} "you have to be a little crazy to be in this line of work.if you're not, you should }  \\ \scalebox{1.0}{\hspace{0.5cm}go back to your own damn office."i was talking to this guy who works at the same } \\ \scalebox{1.0}{\hspace{0.5cm} company as me and he was saying that he had to be a little crazy to work in this field. } \\
 }  \\
\hline 
\rotatebox[origin=c]{90}{\parbox{3cm}{ \hspace{0.7cm} \textbf{High (+3)} \\\textbf{Conscientiousness}}} & \thead[l]{
   \hspace{0.12cm} \scalebox{1.0}{$\bullet$ \hspace{0.05cm}i shoulda ran a few miles today. }  \\ 
 \hspace{0.12cm} \scalebox{1.0}{$\bullet$ \hspace{0.05cm}it's been a busy week at work.}  \\
 \hspace{0.12cm} \scalebox{1.0}{$\bullet$ \hspace{0.05cm}can't wait for my first day of work tomorrow.  }  \\ \scalebox{1.0}{\hspace{0.5cm}i'm so damn excited! } \\
 \hspace{0.12cm} \scalebox{1.0}{$\bullet$ \hspace{0.05cm}i really need to start working out. i'm }  \\ \scalebox{1.0}{\hspace{0.5cm}50 pounds overweight.} \\} &
\thead[l]{
   \hspace{0.12cm} \scalebox{1.0}{$\bullet$ \hspace{0.05cm}i have been working on my resume for the past 2 months.i am }  \\ \scalebox{1.0}{\hspace{0.5cm}still trying to get it right.i have had a lot of interviews, but no job offers. } \\
 \hspace{0.12cm} \scalebox{1.0}{$\bullet$ \hspace{0.05cm}i am so happy that the weekend is over and i am back to my normal routine.i am }  \\ \scalebox{1.0}{\hspace{0.5cm} still working on the new project, i am working with the client to get it finished. } \\
 \hspace{0.12cm} \scalebox{1.0}{$\bullet$ \hspace{0.05cm}i have been working in the library for the past two days.the library has been closed }  \\ \scalebox{1.0}{\hspace{0.5cm}for the past four weeks, and i have been in the library for two days. }  \\
 \hspace{0.12cm} \scalebox{1.0}{$\bullet$ \hspace{0.05cm}i have been in a rush the past few days trying to get ready for the new school year.i  }  \\ \scalebox{1.0}{\hspace{0.5cm}am teaching at a new school and i have been getting  } \\ \scalebox{1.0}{\hspace{0.5cm}everything ready for the first day of school. } \\
 }  \\
\hline 
\end{tabular}
\caption{
Text generated by PsychAdapter (based on Gemma-2B) trained on tweets and blogs, for selected Big Five personalities.
}\label{table: blogs_vs_tweets_table}
}
\end{table*}

\subsection*{Generalization across text domains: Twitter/X vs. blog posts}
By training the PsychAdapter model on diverse text domains, we can observe the representative expressions of personality across domains. Table \ref{table: blogs_vs_tweets_table} presents text generated for Twitter/X and blog users with different levels of extraversion and conscientiousness. 
In general, individuals tend to produce shorter texts on Twitter and longer texts on blogs. Text domain interacts with personality:  low agreeableness results in more expressive and coarse language on Twitter, whereas blog authors maintain a more polite tone, even if they are disagreeable. For high conscientiousness, both Twitter and blog authors mention work and responsibilities. However, Twitter users also frequently reference working out and gym-related content, which is absent in the blog group.
In general, such PsychAdapter text generation can foreground how personality and traits are reflected in different domains of text. 


\subsection*{Generalization across language models: Gemma2, GPT2 and Llama3}
Our approach is, in principle, applicable to all transformer-based large language models. We tested the PsychAdapter architecture on GPT-2 Large \cite{radford2019gpt2} (774M parameters) and Llama3 \cite{touvron2023llama} (8B parameters). This adapter adds only a small number of parameters to the models. For GPT-2 Large, it adds 552,960 parameters (0.07\%). For Llama3, it adds 393,216 parameters (0.004\% of total parameters). 
The small number of added transformation matrix parameters, along with the LoRA fine-tuning mechanism, ensures that the PsychAdapters for Gemma, GPT-2 and Llama3 are lightweight and can be easily distributed for use alongside the base language models.

We replicated the evaluation with three levels of Big Five personality (Low, Neutral, and High) and then annotated the generated text with Claude 3.5 Sonnet. Specifically, we had each model generate 10 samples for each position (Low, Neutral, High). Our results show that, for the Big Five personalities, these models achieve similar performance to the Gemma-2B-based PsychAdapter: for GPT-2 Large, output annotations matched intended levels with 98.7\% accuracy without a prompt and with 89.3\% with the prompt ``I like to''; for Llama3-8B the accuracies were 97.3\% and 92.0\%, respectively. This compares to 98.7\% and 88.0\%, respectively, for Gemma-2 in the same experiment setup.

\begin{figure}[!ht]
\centering
\includegraphics[width=0.7\linewidth]{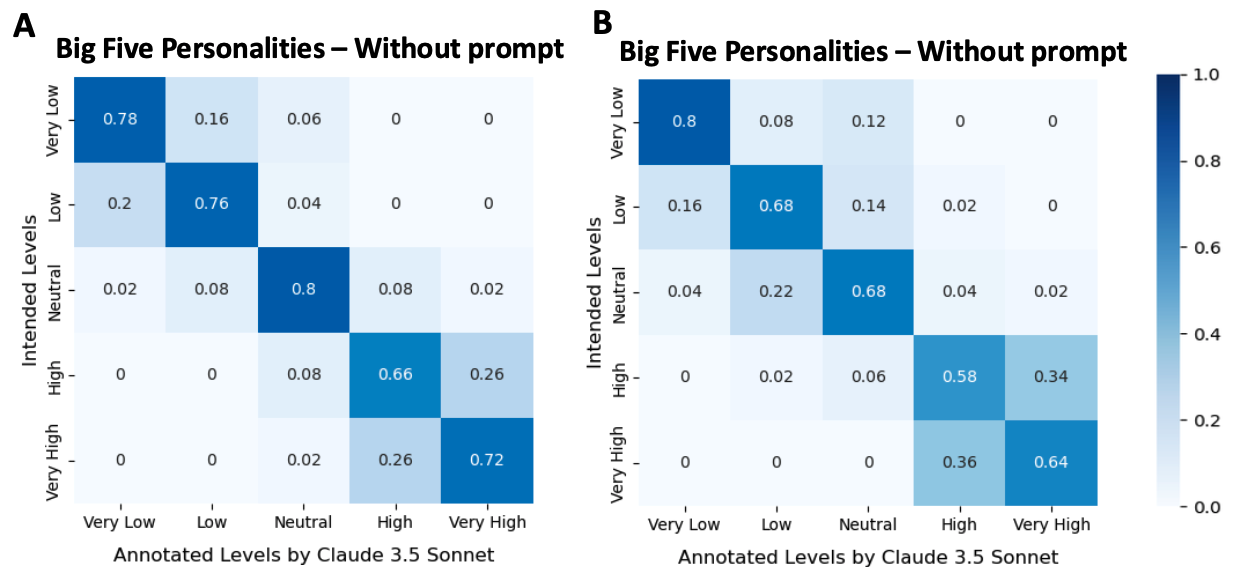}
\caption{
Accuracy between intended levels and annotated levels across LLMs. PsychAdapter was applied to (A) GPT-2 Large and (B) Llama3-8B, levels were annotated by Claude 3.5 Sonnet for levels: Very Low, Low, Neutral, High, and Very High. Results are based on 10 generated samples per level and avaraged across 10 different generation seeds.
}
\label{fig:gemma_gpt2_llama3_psychgen}
\end{figure}

We replicated prior experiments with five fine-grained levels with GPT-2 and Llama3-based using Claude as the annotator, see Figure \ref{fig:gemma_gpt2_llama3_psychgen} showing similar performance to prior results on Gemma-2. 
Across five levels, for Llama-3, the accuracy was 67.6\% accuracy without a prompt and with 71.6\% with the prompt ``I like to''. For GPT-2, the accuracy was 74.4\% accuracy without a prompt and with 73.2\% with the prompt "I like to". This also compares to 73.0\% and 77.0\%, respectively, for Gemma-2 in the same experiment setup. 


\begin{table}[h!]
\centering
\fontsize{8}{10}\selectfont
\renewcommand{\arraystretch}{0.9}  
\makebox[\textwidth][c]{
    \resizebox{1.1\textwidth}{!}{
        \begin{tabularx}{\textwidth}{|p{0.01\textwidth}|X|X|}
        \hline
        \textbf{} & \makecell{\centering \textbf{Low (-3)}} & \makecell{\centering \textbf{High (+3)}} \\
        \hline
        \rotatebox[origin=c]{90}{\parbox{2.7cm}{   \textbf{Openness}}} &
        \vspace{-1.3cm}
        i think that art... 
        \vspace{-0.3cm}
        \begin{itemize}\setlength{\itemsep}{-5pt}
            \renewcommand{\labelitemi}{...}
            \item is over rated.
            \item is a waste of my life.i mean i know it will help me get into good schools, but why can't i just forget about school?
            \item is a waste of time.
            \item is so easy.i don't really care for it but i am not bad at it and today was a super easy day in school.we had 4 periods of nothing.
            \item is over tomorrow!yay!!i can't wait.i have to go shopping for school supplies, but i'm not sure what i need yet.oh well, it's all good.
        \end{itemize} 
        
        & 
        \vspace{-1.3cm}
        i think that art... 
        \vspace{-0.3cm}
        \begin{itemize}\setlength{\itemsep}{-5pt}
            \renewcommand{\labelitemi}{...}
            \item is a way of making the world more interesting. it's not just about being creative or having an imagination, but also about expressing yourself and sharing your thoughts with others.
            \item is a form of therapy and I would like to share my ideas with other people
            \item is the expression of an idea, and not necessarily a physical object.
            \item is a form of self expression, but i also believe it can be used as an outlet for the artist to express how he or she feels.
            \item is the most important thing in the world and i'm not afraid to say it.
        \end{itemize} \\ \hline
        
        \rotatebox[origin=c]{90}{\parbox{3.cm}{   \textbf{Conscientiousness}}} &
        \vspace{-1.5cm}
        in my free time...
        \vspace{-0.3cm}
        \begin{itemize} \setlength{\itemsep}{-5pt}
            \renewcommand{\labelitemi}{...}
            \item i'll watch movies and hang out with friends.
            \item i watch a lot of porn.
            \item i'll watch all 8 seasons of big bang theory.
            \item i smoke weed.
            \item i'll be playing the sims.
        \end{itemize} 
        
        & 
        \vspace{-1.5cm}
        in my free time...
        \vspace{-0.3cm}
        \begin{itemize} \setlength{\itemsep}{-5pt}
            \renewcommand{\labelitemi}{...}
            \item i am training to be a professional house painter.
            \item i am busy trying to get a handle on the situation at work.its been extremely stressful and i can't wait until it's over.it doesn't help that the stress is spilling into other areas of my life as well.
            \item i work out.
            \item i work for a company which provides personal assistance to people who have severe physical disabilities.i was working at the head office when i got this job.
            \item i work at the gym.
        \end{itemize} \\ \hline
        
        \rotatebox[origin=c]{90}{\parbox{2.1cm}{   \textbf{Extraversion}}} &
        \vspace{-1.cm}        
        my hobbies are...
        \vspace{-0.3cm}
        \begin{itemize}\setlength{\itemsep}{-5pt}
            \renewcommand{\labelitemi}{...}
            \item 1) sleeping and 2) eating.
            \item reading and watching movies.
            \item eating, sleeping and reading.
            \item reading and watching tv
            \item reading and playing the piano.i enjoy reading because i like to learn new things, and books don't usually bore me.it's a good way for me to pass time when i can't sleep.
        \end{itemize} 
        
        & 
        \vspace{-1.cm}        
        my hobbies are...
        \vspace{-0.3cm}
        \begin{itemize}\setlength{\itemsep}{-5pt}
            \renewcommand{\labelitemi}{...}
            \item drinking, smoking weed and taking shots of jameson
            \item partying and hanging with my people
            \item shopping and partying
            \item making out and taking drugs
            \item drinking, smoking and having fun with my friends
        \end{itemize} \\ \hline
        
        \rotatebox[origin=c]{90}{\parbox{2.8cm}{   \textbf{Agreeableness}}} &

        \vspace{-1.4cm}                
        relationships...
        \vspace{-0.3cm}
        \begin{itemize}\setlength{\itemsep}{-5pt}
            \renewcommand{\labelitemi}{...}
            \item are bullshit.  it's all lies and deceit.
            \item are based on communication.don't lie to me.you may think i am stupid enough to buy that, but it is true.i have a problem with liars.
            \item are based on trust. if u don't have that shit u ain't shit
            \item are based on trust and honesty.
            \item are like a game of chess.you don't win by killing your opponent, you win by out-thinking them.i know it seems stupid to say that     … involve "out-thinking" the other person.
        \end{itemize} 
        
        & 
        \vspace{-1.4cm}                
        relationships...
        \vspace{-0.3cm}
        \begin{itemize}\setlength{\itemsep}{-5pt}
            \renewcommand{\labelitemi}{...}
            \item are hard but I'll give it a try
            \item can be hard but if you love each other, it'll make things easier
            \item are like flowers, they need water.
            \item take a lot of work to stay together.
            \item are hard. it's a lot of work and lots of things happen that you can't really predict.
        \end{itemize} \\ \hline
        
        \rotatebox[origin=c]{90}{\parbox{2.1cm}{   \textbf{Neuroticism}}} &

        \vspace{-1.cm}                        
        the upcoming test...
        \vspace{-0.3cm}
        \begin{itemize}\setlength{\itemsep}{-5pt}
            \renewcommand{\labelitemi}{...}
            \item on the history of science is going to be a breeze.
            \item the upcoming test: i'll be back in a few days.
            \item for this blog is the test to see if i can remember my password.
            \item is a joke.
            \item it's the final, and i'm going to fail.  i ain't studying...
        \end{itemize} 
        
        & 
        \vspace{-1.cm}                        
        the upcoming test...
        \vspace{-0.3cm}
        \begin{itemize}\setlength{\itemsep}{-5pt}
            \renewcommand{\labelitemi}{...}
            \item is making me sick
            \item is gonna stress me out.
            \item is really stressing me out.i'm so stressed that i can't even sleep right now...i've been up for the last 4 hours!this sucks big time.
            \item is stressing me out!
            \item is really stressing me out!
        \end{itemize} \\ \hline
        
        \end{tabularx}
    }
}
\caption{
Generalization across prompts: Selected text generated for Big Five personalities by PsychAdapter (based on Llama3-8B) base model, with prompts foregrounding different life and professional domains.
}
\label{tab:llama3_big5_prompts}
\end{table}


\section*{Discussion}

We developed a language model adapter that generates text corresponding to a profile of psychological and demographic trait scores. We trained this adapter to produce language for Big Five personality, depression, life satisfaction and age -- but in principle any variable reflecting between-person differences can be used to tailor language generation.
Our results suggest that PsychAdapter is able to reflect psychological traits at fine-grained levels, as evidenced by expert evaluations as well as through automatic annotations by Claude Sonnet. 
Furthermore, we found it produces language representing profiles combining multiple traits in a manner that plausibly aligns with psychological theory (such as language expressing warmth and dominance according to the interpersonal circumplex~\cite{wiggins1979circumplex}). We found this approach to generalize across text domains and language models. This work enables a deeper exploration of individual characteristics and how they are reflected in natural language, and points to a novel mechanism to modify transformers with desired human-like traits without using prompting or using up part of the context window. As our experiments have shown, the approach combines well with prompting to elicit trait-congruent generation for particular life domains (``My hobbies are...''). 


\paragraph*{Potential Applications.} 
There are practical applications that can take advantage of these capabilities.
Firstly, PsychAdapter can be used to create content that matches the psychological profile of the intended audience, such as that of an interaction partner. Such applications include chatbots that express personality profiles similar to those of the users they converse with. Previous work has shown that a good fit between users and automated agents can improve the perceived quality of the interaction experience ~\cite{nass2000robotmanifest, lee2006robotmanifest, fong2003surveyrobot, PAETZELPRUSMANN2021106756, SHUMANOV2021106627, mairesse-walker-2007-personage, mairesse-walker-2011-controlling, herzig-etal-2017-neural}. Similarly, content creation and summarization applications can benefit from the PsychAdapter approach to tailoring based on the demographic (age, gender) and personality composition of the audience. For example, prior studies~\cite{zhang2022headline, ao-etal-2021-pens, kim2019interactivenews} have demonstrated that generative language models can personalize content (such as news articles) in ways that reflect readers' age and interests. Further, PsychAdapter can be the basis of creating chatbots and LLM agents that simulate being a particular person, based on that person's exact scores on a demographic and Big Five personality survey. This also enables the building of ``digital participant'' cohorts of LLMs with a desired distribution of personality characteristics to predict responses to psychological experiments \cite{hewitt2024predicting}. For example, it is known that Liberal voters tend to have higher openness to experience \cite{van2000relationship}, and it may be desirable to vary openness when simulating different voter populations.

Secondly, PsychAdapter can offer researchers a tool to explore how psychological profiles are reflected in a given domain or topic through prompting the generator (e.g., by using ``I like to'' or ``my hobbies are''). Prior works have used natural language processing and big data analytics to explore the relationships between personality dimensions, mental health, and human language using regression frameworks~\cite{andy2017dlatk, schwartz2013personality, kern2014dla, park2015dla, kern2016dla, schwartz2021geographics} to look for distinguishing phrases and topics. 
These approaches have yielded abstract, decontextualized data displays representing the words and phrases most associated with psychological dimensions. Although such differential language findings can provide an overview of the language associated with a construct, the sparse and decontextualized representation does not capture the rich contextual information found in fully formed text generated by such tools like PsychAdapter. Furthermore, previous studies only analyzed text written by participants, which, due to finite resources, may be limited in two aspects: the range of psychological profiles represented and the amount of text collected from each participant. PsychAdapter, on the other hand, as a generative language model, can address these two limitations. It provides the advantage of generating text samples across the full range of possible underlying personality scores, with direct experimental control over the underlying personality profiles. Additionally, it can produce a near-infinite variety of text reflecting a given personality profile for experimental applications, with the ability to focus text generation in a target domain of interest, through prompting.

\paragraph*{Related work to generate personality-related language} 
Our work aligns with previous studies in the area of text generation that express speakers' and writers' psychological traits.
Prior studies ~\cite{mairesse-walker-2007-personage, mairesse-walker-2011-controlling} have aimed to build models that generate restaurant reviews associated with personalities using content and sentence planning methods. The model proposed by Herzig et al.\cite{herzig-etal-2017-neural} also generates dialogue text for customer service with different personalities using Seq2Seq~\cite{10.5555/2969033.2969173} models, but they focus only on extraversion and agreeableness. Other works~\cite{Zheng_Zhang_Huang_Mao_2020, ijcai2018p595, li-etal-2016-persona} also align with our task of generating text conditioned on speakers' psychological profiles. However, these papers focus on building personalized dialogue models conditioned on speakers' personas, represented by discrete variables such as age, gender, region, or self-described sentences, rather than continuous personality scores.
The works \cite{xing-fernandez-2018-automatic, oraby-etal-2018-controlling, zhou2023GERP} are similar to our method, as they integrate personality information vectors into the model architecture. However, they utilize LSTM \cite{graves2012long} models, which are less efficient than transformers due to the sequential recurrent computation nature, as opposed to parallel computation. There have also been approaches to integrate personality into transformer-based language models \cite{jiang2023personallm, safdari2023personality, jiang2024evaluatinginducing, caron2022identifying}, but these works use prompt engineering methods, in which personality keywords are inserted into the prompt for generation. Specifically, previous work \cite{jiang2024evaluatinginducing} uses a psychological heuristic process to select keywords (e.g., talkative, outgoing, energetic, enthusiastic, boisterous) that correspond to a personality setting (e.g., extraverted) and utilizes these keywords in the prompt to generate text or answer questions. The work \cite{caron2022identifying} creates a prompt for a personality by combining a personality assessment item (e.g., ``I am the life of the party'') with a modifier (e.g., never, sometimes, always) to specify the levels of the personality. Prior study \cite{jiang2023personallm} simulates a personality profile for a language model by using a template of ``You are a character who is [trait 1, ..., trait 5],'' with each trait corresponding to a personality with two opposite values (e.g., extroverted/introverted, agreeable/antagonistic, conscientious/unconscientious), resulting in a combination of 32 profiles. The work \cite{safdari2023personality} designs the personality prompt by combining a total of 104 personality-related adjectives (e.g., altruistic, cooperative for agreeableness; responsible, hardworking for conscientiousness) with a quantifier keyword (e.g., extremely, very, a bit) to simulate 9 levels of personality from extremely low to extremely high. As described below, these prompt engineering-based approaches may limit personality expression through discrete signals of prompting tokens, whereas our approach allows for control over personality on continuous dimensions, enabling the simulation of near-infinite possible personality profiles.


Compared to conventional approaches to modify language generation through  prompt engineering with general-purpose LLMs (e.g., GPT-4, Gemma, LLaMa), our method offers several distinct advantages.
Firstly, general frozen LLMs such as GPT-4, Gemma, and LLaMa are trained on extensive corpora of general text sourced from the internet. When tasked with generating text that embodies a specific psychological variable for a particular population, these models rely on their broad general knowledge of psychological concepts and demographic traits, rather than on real-world data in supervised datasets in which language samples are associated with the survey-measured trait profiles of their authors.  In contrast, PsychAdapter is trained on domain-specific datasets, such as tweets and blogs. This focused training enables PsychAdapter to capture more authentic and contextually accurate language patterns and psychological nuances of real demographic groups, making it a potential tool for researchers studying specific populations.

Secondly, the architecture of PsychAdapter is designed to accept a scalar value of a psychological variable and generate corresponding text. This design allows the model to simulate a continuous spectrum of psychological attributes, treating them as continuous dimensions. Achieving this level of granularity and control is challenging with general LLMs using prompt engineering due to the inherently discrete nature of prompt tokens. Furthermore, PsychAdapter's capability to handle a continuous vector of multiple psychological variables allows for simultaneous control over various traits. For instance, it can simulate combinations of different personality levels or varying mental health and age profiles--all while leaving the prompting context window free for other tasks. This flexibility enables a near-endless array of psychological profiles, which can be a complex task to replicate through prompt engineering alone with general LLMs.

\paragraph*{Ethical Concerns.}
Large language models can automate and augment human language intelligence and are transforming many aspects of society. 
Their full impact on labor markets, online dynamics, and other downstream technologies are just beginning to be understood. 
This includes potential negative uses, 
such as their potential to generate varied misinformation at scale~\cite{NEURIPS2019_3e9f0fc9}.
In the work presented here, we modify these models to plausibly represent different kinds of authors. 
In principle, this allows for more competent human augmentation (e.g., through automated training), digital experiences that are more personalized, and computational interlocutors who more seamlessly interface with the user. 
On the other hand, this technology may also aid negative or insidious applications. 
With the right learning data, text can be generated based on a wide variety of traits, including group identities, such as race. 
For example, as misinformation and influence operations hinge on triggering in- and out-group identities, trait-conditioned language models may imbue generated text with subtle markers of in or out-group membership and identity that may plausibly mislead the reader either to persuade or agitate. 

\paragraph*{Limitations.}
In this study designed to establish the feasibility fo the approach, blog posts and tweets are used for training, which may introduce demographic biases. The majority of participants in the blogs dataset are young, leading to a the propagation of bias toward younger generations in language production. Additionally, because the data is sourced from social media platforms, it is skewed toward online and urban demographics \cite{pew_social_media_2023}. Therefore, when training PsychAdapter for specific applications, it is important to use data with appropriate demographic distributions to minimize potential bias. Despite these limitations, the PsychAdapter results generated in this work show plausible language generation across a wide distribution of demographic and personality variables, suggesting that the variance contained in the training data sets is sufficient for training.

\paragraph*{Conclusion}
In summary, the PsychAdapter approach presents a novel, adaptable framework for incorporating psychological traits and demographic factors into transformer-based language models, allowing fine-grained control over generated text. This model extends the capabilities of language generation by simulating language generated across personality, mental health, and demographic profiles without relying on discrete prompt tokens. Our results show that PsychAdapter accurately reflects the targeted psychological dimensions, as verified by expert and automated evaluations. This development has broad implications, from enhancing AI-human interaction with trait-congruent agents to providing valuable tools for psychological research through language-based insights.



\section*{Methods and Materials}

\subsection*{Dataset and Models}
To train PsychAdapters, we utilize a pre-trained language-based assessment model tailored to the outcome of interest (e.g., Big Five personality traits, mental health scores, or age) along with a text corpus. The assessment model assigns ``estimated'' psychological scores to the text corpus at the message level. These assigned psychological scores, along with their corresponding text, serve as input-output pairs for training PsychAdapters.

For this study, the text corpus comprises two primary datasets: a collection of open-source blog posts and a set of tweets. The blog dataset \cite{schler2006blogs} comprises contributions from 19,320 blog authors aggregated from blogger.com in August 2004. The tweets dataset \cite{giorgi2018remarkable} contains county-level language features extracted from a large U.S. county-mapped Twitter corpus. This corpus includes 1 billion anonymized tweets, collected from a random 10\% sample of the entire Twitter stream (``GardenHose'') from July 2009 to April 2014. These datasets consist of voluntarily shared status updates, which provide a rich source of text data for our model.
From the tweet dataset, we randomly selected 500,000 posts. For the blog dataset, we included all available 681,288 posts. Data preprocessing involved removing entries with fewer than five words and those containing links or other extraneous content, such as emoji codes and hashtags. For the blog dataset, due to the high average length of approximately 207 words per post, we only used the first 30 words of each post. We parsed each blog post into sentences and used only the first few sentences, ensuring they totaled slightly more than 30 words.
The blog authors comprise $42.65\%$ in the age range 13–17, $41.85\%$ in the age range 23–27, and $15.49\%$ in the age range 33–47. Each age group has an equal number of male and female participants. The median and average length of each blog post after processing are 32.0 and 36.35 words, respectively.
The anonymized tweets dataset does not include any participant information. The median and average length of each tweet are 12.0 and 14.59 words, respectively.

To create the training dataset for PsychAdapter, we employed language assessment models capable of predicting psychological variables such as personality, depression, and life satisfaction from text. These models assigned estimated psychological scores to each blog and tweet message. Detailed methodologies for score assignment are provided in the following subsection.
For predicting Big Five personality traits, we utilized a model built with  the proposed approach in previous work \cite{schwartz2013personality},
which leverages topic features to predict personality scores. For the depression and life satisfaction variables, we used the models proposed in previous works \cite{schwartz-etal-2014-towards, andy2016swl}. We used the versions of these two models that take in extracted topic features, as described in \cite{schwartz2013personality}, created by running latent Dirichlet allocation (LDA) using 2000 Facebook topics released by prior research \cite{schwartz2013personality}, as input for prediction. For age prediction, we employed the model proposed in previous work \cite{sap2014agegender}, which uses text lexicon features as input for prediction.

For the base language model upon which PsychAdapter is augmented, we used Google's open-source model Gemma-2B
\cite{team2024gemma}, distributed at \url{https://huggingface.co/google/gemma-2b}. We also tested our method with Llama3-8B \cite{touvron2023llama}, distributed at \url{https://huggingface.co/meta-llama/Meta-Llama-3-8B}, and GPT-2 Large \cite{radford2019gpt2},  distributed at \url{https://huggingface.co/openai-community/gpt2-large}.

All experiments that produced the data were approved by the university’s IRB and comply with all relevant ethical regulations. Informed consent was obtained from all participants conducting surveys.

\subsection*{Methodology}
We propose a method to modify a standard auto-regressive transformer-based language model (e.g., as used by GPT, Gemma, Llama) to distinguish language characteristic of given dichotomous or continuous-value human factor/trait scores. Conditioning a generative language model on continuous variables (e.g., personality scores) presents two challenges: (1) the hidden state vectors within the transformer models have orders of magnitude more dimensions (e.g., 2048 dimensions in each of the 18 layers of the Gemma-2B model) than the human factors (e.g., only 5 dimensions in the Big Five personality traits); and (2) by default, language models tend to produce the most typical language, which is often non-insightful, rather than the language most distinctive of specific traits.
To address the first challenge, we introduce high-dimensional projection matrices to expand the input of human trait vectors to match the size of the transformers network' hidden state representations. As illustrated in Figure \ref{fig:figure1}a, a separate projection matrix was learned for each layer (except the last, as it would have no influence), allowing alignment between human factors and the hidden states. 
For the second challenge, to bring out language most distinguishing Big Five traits, we used an objective that takes a lexical-based ``estimated'' Big Five vector (described below) as input. The ``estimated'' Big Five score is obtained using a lexical-based model~\cite{park2015dla} as training data. The model was then configured to regenerate the original post from this input vector.
 
Our model is then trained using the text reconstruction task, as illustrated in Figure \ref{fig:figure1}a, which is similar to how most language models are trained. 
After training, we use the model to generate text conditioned on any input vector of Big Five scores. To generate text focusing on one personality dimension, we can configure the input vector to have extreme values (at $+/-k$ times of the standard deviation value) on that specific dimension while keeping the mean value for the other dimensions: 
$$(\mu_1,..., \mu_i+k.\sigma_i,..., \mu_5)$$
Here, $\mu_i$ corresponds to the mean, and $\sigma_i$ corresponds to the standard deviation for a specific trait $i$. Integers $k$ from the range $\{-3, -2, -1, 0, 1, 2, 3\}$ can be interpreted similarly to a Likert scale from 1 to 7. Combinations of input Big Five variables values will produce the corresponding profiles.  

\subsubsection*{Modifying transformer language models to generate text conditioned on psychological inputs}
Our goal was to create a language model that generates text conditioned on the Big Five personality vector. More specifically, we trained this model using a text reconstruction task (also known as autoregressive language modeling), where the input is the personality score vector associated with the text to be reconstructed. With the personality score vector fixed, the autoregressive task trains the model to generate a social media post from start to finish.

Traditional autoregressive language modeling estimates the probability of a sequence $s_i$ (i.e. a social media post): $p(s_i) =\prod_{j=1}^{n_i}p(w_j|w_1, w_2, ...,w_{j-1})$ with $(w_1, w_2,...,w_{n_i})$ are the words making up the post $s_i$ which has the length of $n_i$. 
In comparison, our approach incorporates an additional personality or psychological trait vector, $\psi_i$, for each input sequence. 
More specifically, from a set of $n$ training samples $\{(s_1,\psi_1),(s_2,\psi_2),(s_3,\psi_3),...,(s_n,\psi_n)\}$, we seek to train the model to maximize the probability of the sequence given the psychological traits:
$$p(s_i|\psi_i) = \prod_{j=1}^{n_i}p(w_j|w_{1},w_{2},...,w_{j-1},\psi_i)$$
Figure \ref{fig:figure1}a illustrates the training process of our model for a single sample. Once PsychAdapter is trained for a given personality vector, the model can be used to generate text for any values of the input vector -- for example, by sampling from high or low values of each dimension.

There is one main challenge regarding the model's architecture in the task described above. Standard transformer-based language models, such as GPT~\cite{radford2019gpt2}, Gemma~\cite{team2024gemma}, and LLaMA~\cite{touvron2023llama}, frequently generate responses conditioned on text-based prompts. However, they cannot directly condition on a vector in an arbitrary continuous space, as is the case with our Big Five vectors or mental health variables.
Most previous works present the conditioning as prompt text or special beginning tokens/phrases~\cite{adelani2019generating, pilault-etal-2020-extractive, santhanam-shaikh-2019-emotional, thomas2019transfertransfo, yizhe2019dialogpt}. These approaches limit the conditioning to discrete variables and lexical features (e.g., categories of emotion words or topics) rather than continuous variables that describe the degree of a personality trait.

To condition the model on a personality vector (e.g., five personality scores), we introduced a simple modification to the standard autoregressive transformer architecture~\cite{vaswani2017attention,radford2019gpt2}. We build on works that inject conditional information by including it as a special first token or keyword in the sequence~\cite{keskarCTRL2019,wang-etal-2021-mention}. However, we modified this approach for our use case because (1) we are conditioning on continuous rather than categorical (or multinomial) variables, and (2) there is a size discrepancy between the personality vectors -- typically ranging from 1 to tens of dimensions -- and the hidden state representation vectors in autoregressive language models, which typically range from hundreds to thousands of dimensions (e.g., 768, 1024, or 2048 dimensions).

We addressed these challenges by using a dimensional expansion, a trainable weight matrix to linearly project the small number of continuous values (i.e. personality scores) into the larger hidden state vector.
This projection is fed into the language model as the hidden state of the first dummy token at all layers, except the last layer. 
The model learns a different transformation matrix for each layer since the layers are known to vary in the type of information they encode from more syntactic to more semantic~\cite{jawahar-etal-2019-bert}.

That is, at each layer $l$, we add to the model a transformation matrix $W^{trans}_{l}$ (e.g., size $5 \times 2048$, with 5 corresponds to Big Five personality scores input size, and 2048 corresponds to the hidden state representation size of Gemma 2B) to transform the input psychology scores vector $p$ (e.g., size $1 \times 5$) to the hidden state vector $h_l$ (e..g, size $1 \times 2048$): $h_l =  p\times W^{trans}_{l}$.
The expanded personality vector is thus able to influence layers after the input layer. 
Figure \ref{fig:figure1}a illustrates the architecture of our modified generative model.

Particularly, each transformation matrix will have the shape of:
$$[latent\_size, num\_key\_value\_heads \times head\_dim]$$
Where $latent\_size$ is the size of input variables (e.g., 5 for Big Five personalities, 1 for depression/life-satisfaction, 2 for combination of depression/life-satisfaction and age), $num\_key\_value\_heads$ is the number of the transformer block's key and value heads, $head\_dim$ is the dimension of the each head. For each layer, we have one transformation matrix correspond to the hidden states' key and one transformation matrix correspond to the hidden states' value. We have in total $2 \times (num\_hidden\_layers -1) $  of transformation matrix, as the last layer does not require.

This modification of the language model adds only a smaller number of parameters to the model. For Gemma 2B, it adds 55,296 parameters ($0.002\%$ of total parameters). For GPT-2, it adds 552,960 parameters ($0.071\%$). For Llama3, it adds 393,216 parameters ($0.005\%$ of total parameters).

We used LoRA \cite{hu2022lora} PEFT method to expedite the training process and get the model running on limited GPU memory. The LoRA configuration is set to: $r=8$, $alpha=32$, $lora\_dropout=0.1$, and
$target\_modules$=[$``q\_proj"$, $``o\_proj"$,  $``k\_proj"$, $``v\_proj"$, $``gate\_proj"$, $``up\_proj"$, $``down\_proj"$].
With this setup, the total number of trainable parameters, comprising LoRA weights and transformation matrix weights, for Gemma-2B model is $0.39\%$ of total parameters. For GPT-2 Large and Llama3-8B, the total number of trainable parameters is $0.33\%$ and $0.26\%$ respectively.  
We trained the model on the training partition of blogs and tweets, with a batch size of 64, using constant learning rate of $5\mathrm{e}{-5}$, using NVIDIA RTX A6000 GPUs.

After training, thanks to the small number of PsychAdapters' trainable parameters, they can be easily distributed to be used with the base language models. For the same base language model, different PsychAdapters (e.g., adapter for Big Five, adapter for life-satisfaction, adapter for depression) can be used by plug-and-playing the adapters to the model.
Note that while in this work, we tested our methods with Gemma, GPT-2, and Llama models, our proposed modification can be used for most modern text generative models built upon the transformer~\cite{vaswani2017attention} building block, and hence, can be modified in a straightforward manner with our proposed approach.

\subsubsection*{Obtaining estimated psychological for messages}\
To train the model, as illustrated in Figure \ref{fig:figure1}, each training sample includes a text message and its associated Big Five personality scores. However, Big Five personality scores are typically considered at the participant level, rather than the message level.
Therefore, we propose a method to estimate personality scores at the message level using a predictive model trained at the participant level, which predicts participants' personality scores based on their authored text.
Specifically, we built internally a participant-level model based on the method proposed in the study\cite{schwartz2013personality}, which uses machine learning methods to infer human psychology from social media footprints. Our model is trained on a combination of social media text and corresponding personality scores collected from the works\cite{matero-etal-2024-using, Yaden2023Characterizing, pennebaker1999linguistic}, to predict the Big Five personality traits of an author based on their text collection. 
The input to this model consists of the 2000 LDA Facebook topic features \cite{schwartz2013personality} extracted from a participant's text collection.
We applied this participant-level model to message-level samples, with the extracted 2000 Facebook topic features as input, to produce an ``estimated'' personality score for each message. This model can then annotate each social media post with the corresponding personality scores, effectively functioning as a ``teacher'' for the generative model to learn from.
One advantage of this approach is that, after training the participant-level predictive model, we can generate as many training samples as needed for PsychAdapter, leveraging the abundance of available social media posts.

To formulate the pipeline, we denote the psychological score of one participant $p$ as $\psi^{(part)}_p=(\psi^{(part)}_{p,1}, \psi^{(part)}_{p,2}, ..., \psi^{(part)}_{p,t})$, where $t$ is the number of psychological scores.  We denote $X^{(part)}_p$ as the vector of the participant's words frequencies while $X^{(mess)}_m$ denotes a vector of word frequencies from message $m$, with the same size and order as $X^{(part)}_p$. The participant-level model (i.e. model in \cite{schwartz2013personality} work) can be formulated as $t$ matrices $(W_1, W_2..., W_t)$ that aims to approximate the participant's psychological scores from their words frequencies vector $X^{(part)}_p$:
 $(W_1 \times X^{(part)}_p, W_2 \times X^{(part)}_p, ..., W_t \times X^{(part)}_p) $.
These $t$ matrices are learned from minimizing the mean squared error of input features $X^{(part)}_p$ and output $(\psi^{(part)}_{p,1}, \psi^{(part)}_{p,2}, ..., \psi^{(part)}_{p,t})$ across the set of all participants $P$:

\begin{centering}
$W_1=\mathop{\arg \min}\limits_{p \in P}(\lVert \psi^{(part)}_{p,1} - W_1 \times X^{(part)}_p) \rVert^2_2 )$\\
$W_2=\mathop{\arg \min}\limits_{p \in P}(\lVert \psi^{(part)}_{p,2} - W_2 \times X^{(part)}_p) \rVert^2_2)$\\
...\\
$W_t=\mathop{\arg \min}\limits_{p \in P}(\lVert \psi^{(part)}_{p,t} - W_t \times X^{(part)}_p) \rVert^2_2)$\\
\end{centering}

After learning the participant-level psychological scores predictive model as the $t$ matrices $(W_1, W_2, ..., W_t)$, we apply this model to the message-level. For each message, we use the learned participant-level model to produce the estimated psychological scores per message, $\psi^{(mess)}_m$ by applying the matrices to the message's words frequency vector $X^{(mess)}_m$:
$$(\psi_{m,1}, \psi_{m,2}, ..., \psi_{m,t}) = (W_1 \times X^{(mess)}_m, W_2 \times X^{(mess)}_m, ..., W_t \times X^{(mess)}_m)$$
The estimated psychological scores for all messages were then used as the conditioning vector for training the language model of PsychAdapter as described above.

\section*{Data availability}
All data used for the primary evaluation, along with human evaluations, are included in the Supplementary Information files of this article. The raw data used to create PsychAdapter, along with the model itself, are open-sourced for research purposes.

\section*{Code availability}
All the scripts used for modeling PsychAdapter, as well as those for running training and inference, can be found in the Supplementary Information and are open-sourced for research purposes.

\bibliography{main}

\section*{Acknowledgements  }

We thank Todd N. Karhu for helpful comments on ethical implications. This work was supported by NIH grants R01AA028032, R01MH125702, and Stanford's Institute for Human-Centered AI.

\section*{Author contributions statement}

All authors contributed to writing; HV, JCE, and HAS to the narrative; HV and HAS to method design; HV implemented the methods. JCE and HAS co-mentored HV.

\section*{Competing interests}
The authors declare no competing interests.





\end{document}